\documentclass{article}

\usepackage[utf8]{inputenc} 
\usepackage[T1]{fontenc}
\usepackage{amsmath, amssymb}
\usepackage{graphicx}
\usepackage{subcaption}
\usepackage{appendix} 
\usepackage{hyperref} 
\usepackage{textcomp}

\title{The Endless Tuning\\[0.5em]
\large\textit{An Artificial Intelligence Design To Avoid Human Replacement and Trace Back Responsibilities}}

\author{
  Elio Grande\\
  University of Pisa, Department of Computer Science\\
  \texttt{elio.grande@phd.unipi.it}
}

\date{Preprint – December 2025}  

\begin{document}

\maketitle\footnotetext{
This paper introduces the “Endless Tuning” framework, developed by the author.  
A code demonstration, including subrepositories for each case study, is available at:  
\url{https://github.com/eliogrande/TheEndlessTuning.git}.

Interactive versions can also be executed via Google Colab:  
\url{https://colab.research.google.com/drive/1m1mDWTlE5egzT_oSR18hHLlcwwrX401N?authuser=1\#scrollTo=3pMgbEeXeB5X}.

This material is released under a Creative Commons Attribution-NonCommercial-ShareAlike 4.0 International License (CC BY-NC-SA 4.0).  
Commercial applications (including integration into proprietary systems or consulting services) require prior written consent from the author.
}

\begin{abstract}
The \textit{Endless Tuning} is a design method for a reliable deployment of artificial intelligence based on a double mirroring process, which pursues both the goals of avoiding human replacement and filling the so-called responsibility gap \cite{matthias2004responsibility}. Originally depicted in \cite{fabris2024towards} and ensuing the relational approach urged therein, it was then actualized in a protocol, implemented in three prototypical applications regarding decision-making processes (respectively: loan granting, pneumonia diagnosis, and art style recognition) and tested with such as many domain experts. Step by step illustrating the protocol, giving insights concretely showing \textquotedblleft a different voice\textquotedblright \cite{gilligan1993different} in the ethics of artificial intelligence, a philosophical account of technical choices (\textit{e.g.}, a reversed and hermeneutic deployment of XAI algorithms) will be provided in the present study together with the results of the experiments, focusing on user experience rather than statistical accuracy. Even thoroughly employing deep learning models, full control was perceived by the interviewees in the decision-making setting, while it appeared that a bridge can be built between accountability and liability in case of damage.
\end{abstract}

\vspace{1em}

\section{Introduction}
\label{sec:chap1}

Fred, a 10 years old guy, is a patient of a pediatric cancer ward. Since his case is quite a complicated one, both John --- a thirty-year experienced pediatric oncologist --- and an artificial intelligence system are asked of making a diagnosis. Unfortunately, the two predictors reach different conclusions. Moreover, contrary to the prediction of the model, the diagnosis carried out by the physician suggests the need for surgery. Despite Fred's family being very supportive and even available to take such a risk, the hospital's management expresses a negative judgement, not giving the permission to proceed with the surgical operation. Is it that the physician's diagnosis was not convincing? Is it, so to speak, a matter of accuracy? No, and that is why a \textquotedblleft biased'' description was just provided, not sufficiently lingering on the \textquotedblleft expertise\textquotedblright the artificial intelligence model gained during training. Indeed, \guillemotleft should anything go wrong and should we be suited\guillemotright, the administration contends, \guillemotleft we would certainly lose the case: we did not follow the artificial intelligence's prescriptions\guillemotright. Said in passing, John hands his resignation.

Much work has already been carried out regarding over-reliant behaviour: \textit{ex multis}, not to mention the foundational \cite{kahneman2011thinking} on System 1 and System 2, the persistent character and the sluggish one of our mind, \cite{skitka1999does} experimentally recorded commission and omission errors due to cognitive laziness in the presence of automated decision aids; a pilot study in \cite{bansal2021does} observed that explanations in artificial intelligence can well increase reliance on recommendations even when they are incorrect; about to predict whether a person’s annual income would exceed \$50K based on some demographic and
job information, \cite{zhang2020effect} noticed that showing confidence scores above 80\% significantly enhanced trust through an increase in switch percentage (changing idea); \cite{buccinca2021trust} successfully tested three cognitive forcing functions (on-demand explanations, updating decisions, waiting time to obtain explanations) in a decision-making process; in a cooperative spirit, machine learning techniques like \textit{learning to reject} \cite{cortes2016learning} and \textit{to defer} \cite{madras2018predict} suggested a socratic artificial intelligence \cite{punzi2023towards}. 

While maintaining a general definition of trust as \guillemotleft the attitude that an
agent will help achieve an individual’s goals in
a situation characterized by uncertainty and
vulnerability\guillemotright \cite{lee2004trust}, what the aforementioned anecdote (which, be it true or not, it suffices to reckon it as plausible) brings to light, even independently of the faithfulness of outcomes, lies instead in the ontological properties attributed to the trustee. Indeed, what is it that really seems to matter far beyond and before the automation bias itself? In a deeper sense, the fact that \textit{we bestow on artificial intelligence the ability to tell something}, adorning with a \guillemotleft style: impressionism\guillemotright, a \guillemotleft trustworthy applicant: yes\guillemotright, or even a \guillemotleft exchange value: \$15k\guillemotright what is simply a feature, often not transparently computed. But if an algorithm can \textit{say} something, it can \textit{affirm} or \textit{deny}; if it can affirm or deny, it can \textit{judge}; if it can judge, it can \textit{think} and has \textit{willingness}. A bit too much, isn't it? 

Strictly coupled with the mandate of action given to an algorithm, the fateful question arises: who is going to pay in case of damage --- obviously given that a causal connection (or at least a reasonable probabilistic one) between the damage itself and the artificial intelligence system is proved? Such a noble concept as the European HLEG's requirement of human agency and oversight \cite{hleg2019ethics}, be it meant as human-in-the-loop (see f.e. \cite{wu2022survey}) or, in reverse, a machine-in-the-loop paradigm (see, under certain respects, \cite{miller2023explainable}), could be not so effective until the so-called responsibility gap \cite{matthias2004responsibility} is filled, \textit{i.e.} until the artificial agent's agency gets \textit{hooked} to human responsibility. Yet, how to achieve this goal? Not surprisingly, brotherhood comes about among the artificial intelligence ethicist, the computer scientist (or the developer) and the jurist (or the lawyer) --- the first one becoming the link between the other two figures. 

Starting by focusing on \guillemotleft what acting means\guillemotright, \cite{fabris2024towards} underlines the absence of awareness and willingness in artificial agency, reporting the distortion produced by some anthropomorphic projection. Nonetheless, stumbling just a step outside the perimeter of this projection, breaching the theoretical sphere and literally staying \textit{out of sight}, machine learning algorithms \textit{act}. Despite, for example, remaining blind and deterministic from a mathematical viewpoint, the neurons of a neural network shooting layer by layer cease to appear so blind with respect to the meaning or, better, the sense of the activations themselves. Again \textit{in the act of activating}. A veracious and tangible way to ethics of artificial intelligence could precisely capitalize on this discrepancy, in the effort to \textit{get in touch}, to establish a \textit{relation} \cite{Buber1970-BUBIAT,fabris2016relazione} between different modes of acting. We put forward, indeed, that a design that deals with this relation, traversed back and forth and covering the various entities involved (the producer, the developer, the interface designer, the end-user, and so on), might well constitute the intersection point between accountability and liability. 

A comprehensive study concerning the criteria of imputability with respect to artificial intelligence and civil liability \cite{dal2024intelligenza} shows that currently, in the European Union's law, a complementary approach is looked for between the \textit{ex ante} and \textit{ex post} dimensions of responsibility, where \guillemotleft to the first one corresponds a goal of ascription of responsibility concerning the subjects that produce and use AI systems in the context of their activities\guillemotright --- the EU Artificial Intelligence Act \cite{EU2024AIAct}, specifying requirements and obligations for providers and operators according to the risk implied toward fundamental human rights --- \guillemotleft while the second deals with completing the protections' framework set up by the safety legislation, providing protection to damaged people also with respect to risks not considered by this last one\guillemotright \cite[p. 321]{dal2024intelligenza}\footnote{Translation is mine.} --- the new Product Liability Directive for defective products \cite{EU2024PLD}, with some space left for national law concerning objective liability. Due to the complexity of the supply chain (known as the ``problem of many hands'' \cite{nissenbaum1996accountability}) and the partially unpredictable conduct of the algorithms themselves, the rules of accountability can serve the purpose of better distributing responsibility \cite[pp. 322 ff.]{dal2024intelligenza}, whereas liability usually focuses more on the entity of the damage to be repaired.

\textit{In medias res}, partly entering and partly building the previously mentioned relation, we propose the \textit{Endless Tuning}, an ethical design method of artificial intelligence that, at least with regard to decision-making processes (classification or regression tasks), allows to employ artificial intelligence models of nearly any depth avoiding the replacement of the user --- but correspondingly requiring them of being competent --- and still having the possibility of tracing back responsibilities. The general method itself, which has already been depicted in \cite{fabris2024towards}, will be given further insights in Section \ref{sec:chap2}. Secondly, a protocol has been built upon this method, which will be presented in Section \ref{sec:chap3}, step by step explaining its modules and the reasons thereof. In the third place, the protocol itself was implemented in three prototypical versions (that is, so as many apps), according to three different tasks (loan granting, art style recognition, and chest x-ray pneumonia diagnosis), and it was tested through interviewing three domain experts (see \ref{AppendixA},\ref{AppendixB} and \ref{AppendixC} for transcriptions). Section \ref{sec:chap4} will provide technical details of the implemented prototypes and sum up the results of the interviews. Section \ref{sec:chap5} will further discuss the calibration of trust and mandate of action and reflect on whether and how the responsibility gap gets filled.

\section{Soft interpreters. Insights on the method}
\label{sec:chap2}

Practical responses should be given to practical issues; but, even within the framework of fundamental rights, not by imposing a specific ethical viewpoint embedded in a model or a system: rather than ethics, that would be moralism. Similarly to a pro-ethical design \cite[p. 190]{floridi2014fourth}, the Endless Tuning does instead provide the operator with the freedom to fairly deploy an artificial intelligence system. Such freedom is not meant to be merely bounded by the properties of the system itself (positively, thanks to the GUI's layout, rather than negatively, by the means of data-driven biases \cite{ntoutsi2020bias}), rather resembling a rail on which actions are forced to run. Concerning the term \textquotedblleft interface'', \cite[p. 7]{zannoni2024design} underlines how the suffix \textquotedblleft inter-'' before the substantive \textquotedblleft face'' indicates properties of commonality, link and relation among human beings and between and with real or virtual artifacts. \guillemotleft It is necessary to reflect on the fact that the interface does not exist and it is just an abstract entity which we refer to [...]\guillemotright\footnote{Translation is mine}. Delving into such an idea,\textit{ an interface would appear to us as the way in which a relation reveals itself}. 

\begin{figure}
    \centering
    \includegraphics[width=0.5\linewidth]{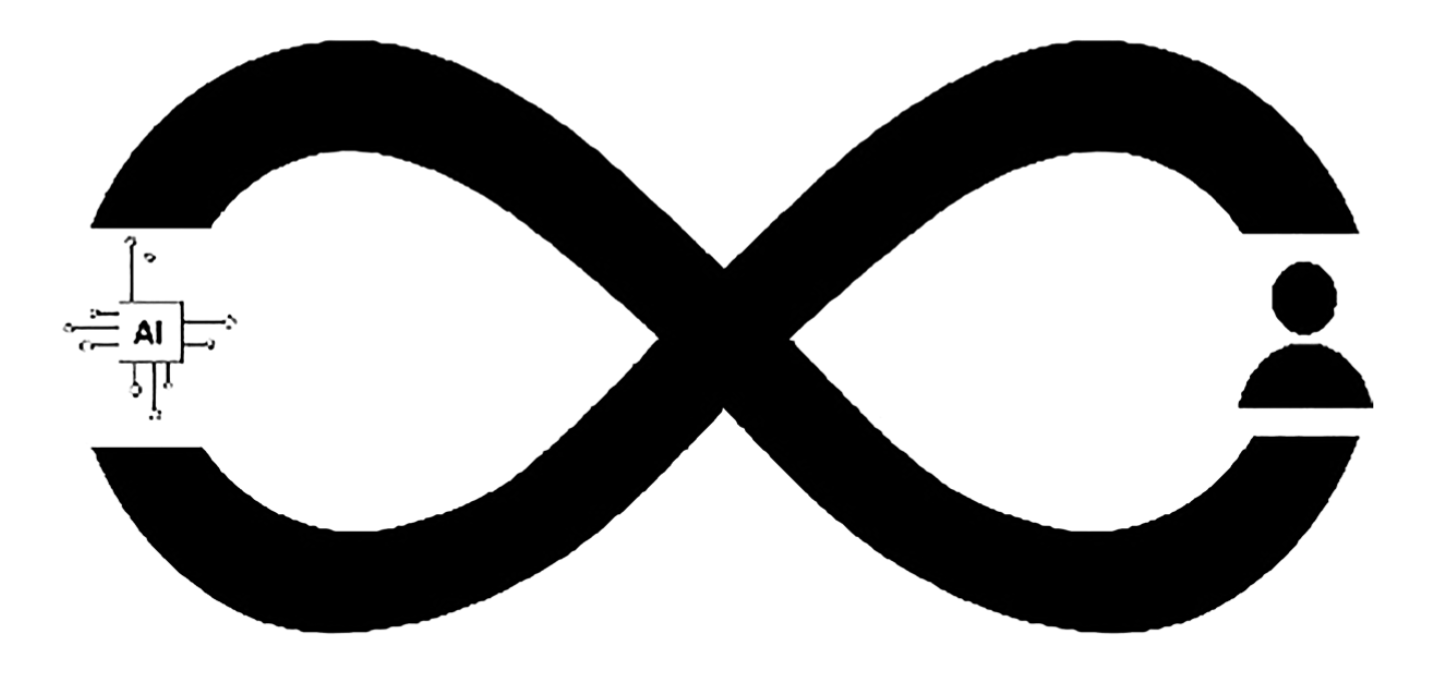}
    \caption{A graphical representation of the double loop of Endless Tuning (image taken from \cite{fabris2024towards}; originally made by the same author of this paper; received permission by the editors).}
    \label{fig:fig1}
\end{figure}

Our method is composed of only two general rules: 

\begin{center}
  \begin{itemize}
    \item \textit{ex ante}, the system should manage to make the operator reflect, while the operator should manage to put learning (and/or, if necessary, explainability) constraints on the system, so as to \textit{tune on each other} (the reference is to radio electronic circuits, where a beat is performed between a carrier and a receiver wave), the system being adaptive and ergonomic (see Figure \ref{fig:fig1} as a visual representation);
    \item some information about every interaction, adjustment (\textit{e.g}., finetuning) and outcome at least partially due to random processes should be permanently recorded, so as that, \textit{ex post}, the decision-making process could be reviewed ``in slow motion''.
  \end{itemize}
\end{center}

A sort of Continual Learning \cite{lesort2020continual} is in fact depicted here, composed of a double negative feedback. Since the present article concentrates on ethical design, concerning proper machine learning challenges and possible objections thereabout not only might it be worth noting that Continual Learning is a growing field, but also that here we are proposing machine learning on single (or few) tasks, carrying out finetuning by the means of simple rehearsal techniques, which can prove themselves effective. Yet, without referring to specific architectures, we would like to go straight to one point. Our issue certainly lies in contrasting each other's biases, but when a human learner/decision-maker comes up beside a machine learning model, such artificial intelligence device emerges as an action's \textit{center of gravity}. Such a human learner, getting involved and having to balance choices \textit{in the meanwhile of interacting}, gives the learning process a \textit{real time} shade, also dealing with an even subtle behavioral deviation margin. The idea to reciprocally profit from each other's fragility was conceptually illustrated in \cite{fabris2024towards} --- after all, the moral, cognitive, physical and even metaphysical fragility of human beings has been notorious since the dawn of time, while for example the media sensation about the COMPAS software \cite{larson2016compas,angwin2016machinebias} publicly disclosed an Achilles heel of artificial intelligence. Sections \ref{sec:chap3} and Section \ref{sec:chap4} will dive into the actual implementation. Yet, a ground point remains to be clarified: \textit{is there really a different voice to be heard?} \cite{gilligan1993different} If machine learning algorithms reflect us as mirrors \cite{fabris2012etica,valera2022espejos}, can they mirror themselves in us? Or is instead the Endless Tuning itself based on a false friend, victim of a silly deceit \cite{natale2022macchine}? 

Despite being of utmost significance, accuracy itself might not be the only key. As a statistical measure, accuracy is strictly dependent on the size and the internal variety of the learning data sample --- a shift in the data distribution might be lethal in inference mode and, as for the training phase, produce a sudden loss in performance called \textit{catastrophic forgetting} (see, f. e., \cite{chen2018continual,lesort2020continual}). Additionally, we usually employ artificial intelligence \textit{one inference at a time}. Moreover, outside the laboratory setting, it is reasonable to find out that people use artificial intelligence precisely when they ignore the ground truth. Still, except for transparent models like decision trees and so on, it remains difficult to clearly explain even a single inference because \guillemotleft internal states of machine learning models are
non-concepts, concepts that have no intuitive equivalent in the real world and that can only be
represented in terms of what they are not\guillemotright \cite{offert2017know}.  

Such ``vulnerability'' could instead become a pivotal resource. Little by little, passing through a spectrum of presence \textit{not} meant as subjectivity, an artificial intelligence algorithm might disclose itself, if not as an interpreter, at least as a useful prompter. Take a vanilla neural network as a guinea pig, not even a deep one, simply trained in supervised fashion. Does it sensibly \textit{emerge} as an entropy reducer, even if such entropy is not computable? Or, if you prefer, \textit{does it return nonsensical truths as ancient sophists distressed Plato} \cite{plato_sophist_1986} \textit{by telling sensible lies?} 

Suppose the model predicts something. Strictly speaking, a feature vector enters the network making neurons consequently ``spike'' in a merely deterministic fashion. Our lack of knowledge about the internal process could nonetheless lead to unforeseen and sometimes enigmatic results. Tricky questions: how artificial is such an incomprehensible artifact? So, why are we ignorant about that? It is not solely a matter of complexity. Let us take a step back. Our model was trained according to a meta-algorithm (backpropagation) and through a \textit{meaningful (for us) target}. It is us, for example, who connect the neuron of the last layer indexed 2 with "dog". Consequently, while its obscurity conceptually increases, it comes out that its function has \textit{some meaning beyond a strictly numerical one}. On the one hand, information gets inscribed in a digital object. Simondon would perhaps have called it a \textit{transductive} process, carried out \guillemotleft by a mechanism that is analogous to that of perception in the living\guillemotright, to build up a transducer  \cite[pp. 155-156]{alma99244708110502171}. On the other hand, not only is this transductive process (loss and gradient calculation, optimizer steps) itself unreachable even recording and reviewing all the steps with naked eye, but no tool exists to catch \textit{how} the backpropagation or the inference treats that other sphere of sense beyond a numerical one. Right within this crevice (which could come out to be a rift), neither able to know, nor to think, nor to tell anything, artificial intelligence acts, somehow beyond categories: \textit{its acting is without action }\cite[pp. 38-40]{fabris2002paradossi}.

On a similar path, Romele \cite[pp. 89 ff.]{romele2019digital} suggests the concepts of ``imaginative machines'' and ``\textit{e}magination'' --- a form of productive imagination which, sometimes obscurely, embeds the schematizations of our mind in a digital device.\footnote{At the time of \cite{fabris2024towards}, I had not had the occasion of reading this book. I would like to mention it now. To my pleasure and surprise, a double loop similar to the one of the Endless Tuning was depicted there (p. 104). Despite some differences, a somehow common spirit drove us to a similar intuition.} Drawing from sociology and hermeneutics, it talks about \guillemotleft a new form of complex mediation, no longer reducible to a single cognitive system\guillemotright \cite[p. 98]{romele2019digital}: the new algorithmic machines, nowadays all imaginative at a certain level of abstraction, taking a break from human imagination. \cite{romele2019digital} mentions \cite[p. 2]{burrell2016machine}: algorithmic opacity above all \guillemotleft stems from the mismatch between mathematical optimization in high-dimensionality characteristic of machine learning and the demands of human-scale reasoning and styles of semantic interpretation\guillemotright. It might not be news to say that our vanilla neural network has or even is a self-standing structure, but it is not at all obvious to suggest it is a sort of \textit{reactive} one. Our goal is to land, say, on the dark side of such opacity: let us move a last step before passing to the protocol.

A crystalline germ containing \guillemotleft no more than an extremely small amount of energy\guillemotright{} gets in touch with an amorphous body endowed with potential energy: \guillemotleft its structure and its orientation take control of this energy of the metastable state\guillemotright{} and the crystallization starts \cite[pp. 80 ff.]{gilbert2020individuation} \textit{Mutatis mutandis}, such crystalline germs remind us of the phenomenon of \textit{adversarial examples}, which in \cite{fabris2024towards} were philosophically suspected to be bugs in the nature of artificial intelligence itself, since an usable object seems to become unusable precisely because of its usability. Adversarial examples are data instances affected by extremely small perturbations (\( x - x' < r \)), which are capable of profoundly (and, paradoxically, precisely) altering the outcome or the explanation (see \textit{e.g.} \cite{razmi2023interpretation}) of a model with respect to the original data. For example, a picture that with naked eye is perfectly recognizable as one of a school bus may well be classified with high confidence as an ostrich. Starting from \cite{szegedy2013intriguing}, discovering them and calling them \guillemotleft somewhat universal\guillemotright, much work has been carried out trying to understand their nature, typology, construction methods and defenses (\textit{ex multis}, \cite{goodfellow2014explaining,yuan2019adversarial,zhang2019adversarial,zhao2017generating,xie2019information,li2021verifying,madry2017towards,moosavi2017universal,kurakin2018adversarial,chattopadhyay2019curse,elsayed2018adversarial,hendrycks2021natural,ilyas2019adversarial,shafahi2018adversarial,buckner2020understanding}). 

Far from seeking solutions to such mathematically complex problems, we are merely gathering insights. While \cite{chattopadhyay2019curse} or \cite{shafahi2018adversarial} focus on a curse of data dimensionality, reporting the ease of moving a point outside the bounds designed by the classifier in a high-dimensionality setting, \cite{ilyas2019adversarial} showed the existence of \textit{useful} but non-robust features --- a standard model trained on a non-robust, mislabeled dataset was nonetheless capable of generalizing on a standard test set. Being useful but imperceptible, such features appeared peculiarly meaningful \textit{for the model}. Yet, \cite{ilyas2019adversarial} attributed the \textquotedblleft adversarial'' issue to a possible misalignment between the inherent geometry of data and the l-norm metric employed to build the adversarial examples. Disagreeing with a strongly geometrical approach, \cite{buckner2020understanding} underlines that adversarial examples could be affected by features that are \guillemotleft neither signal nor random noise\guillemotright, calling them \textquotedblleft artifacts'', or \guillemotleft systematic, reproducible patterns in transformed signals that are created by interactions between our instruments and the world\guillemotright, sort of \textquotedblleft entities of mind'' just like Doppler effects. Be it as it may, such errors exist, say, oscillating between being a formal and a material issue. As reading Romele would suggest, Levinas \cite{levinas1963trace} would perhaps have noticed: 

\begin{center}
\small
\begin{quote}
    \guillemotleft In the presence of the other do we not respond to an \textquotedblleft order'' in which signifyingness remains an irremissible disturbance, an utterly bygone past? Such is the signifyingness of a trace. [...] In a trace the relationship between the signified and the signification is not a correlation, but \textit{unrightness} itself\guillemotright.
\end{quote}
\end{center}

Perhaps we have found some levers to lift a relation, say, beyond accuracy and inaccuracy and somehow outside taxonomies of autonomous \cite{tosic2004towards} and moral machines \cite{moor2006nature}. As Section \ref{sec:chap2} has been focusing on \textquotedblleft the trace of the other'' in machine learning, we refer to Section \ref{sec:chap3} concerning the need for a reverse employment of eXplainable Artificial Intelligence algorithms in a hermeneutic fashion. Despite being very far from admitting personality, dignity or consequently rights in computers, the approach of Endless Tuning calls for the government rather than the strict control of artificial intelligence systems, asymmetrically recognizing reciprocal fragility. If we were to define such a relation, we would borrow from Levinas the expression \textquotedblleft a contact without contact''. All in all, it is a soft tie, despite the amount of computational resources needed and the risks involved. Paradoxically, \textquotedblleft dissecting the (nonexistent) soul'' of machines seems the only way to avoid ending up on the couch of an artificial psychoanalyst.

\section{To the thing itself: the protocol}
\label{sec:chap3}

As the Greek origin of the term \textquotedblleft method'' suggests, the Endless Tuning offers the seed ----- the principle, or the beginning ----- for finding a way out of a labyrinth. Just few rules, as Descartes taught us, can indeed well be sufficient. Naturally, putting it into practice requires the development of a protocol or procedure. We do not claim that the procedure described below is the only possible one. However, it certainly embodies the method when applied to decision-making processes --- the case of generative artificial intelligence is left for future research. In turn, each procedure must be itself instantiated, according to the specific environment of usage, through the selection of appropriate frameworks.

\cite{dellermann2019hybrid} defined Hybrid Intelligence as \guillemotleft the ability to achieve complex goals by combining human and artificial intelligence, thereby reaching superior results to those each of them could have accomplished separately, and continuously improve by learning from each other\guillemotright. \cite{hemmer2021human} noted that using interpretability to increase complementary team performance (CTP) might be a tricky issue: findings may be various. \textit{Ex multis}, experimenting with various conditions across two risk assessment tasks, \cite{green2019principles} found that asking users to first make their own decision led to the highest CTP, surpassing even XAI assistance. \cite{bansal2021does} experimentally found that XAI should be used adaptively, depending on the model’s confidence. In addition, one issue (see f.e. \cite{alufaisan2021does}) is that the user's solo performance is not always superior to that of the model with which they collaborate (although discovering patterns can enhance human performance \cite{lai2020chicago}). This raises an important reflection: granted that both the human and the model should be as accurate as possible, if it is true that an accurate and \textquotedblleft persuasive'' model can lead to good outcomes, what about responsibility for errors? Put differently, if we had to choose, \textit{would it be preferable to pursue higher accuracy with less control over potential negative consequences, or slightly lower accuracy with greater capacity for remediation?} We should also take into account the context in which the task is framed as well as, strictly speaking, the design of the interface itself. Is it useful, for instance, to display the model’s confidence? \cite{zhang2020effect} found that it is; however, users may even \textquotedblleft rebel\textquotedblright{} against the recommendations \cite{dietvorst2015algorithm}. After all, we are dealing with the human being, the quintessential \textquotedblleft polytropos\textquotedblright. 

An application protocol was developed for this study. As for relational approaches and protocols which investigate in a similar direction, we should mention here \cite{cabitza2022open,cabitza2023rams} and above all \cite{campisi2023doctoral} which has been inspiring with regard to some technical solution. Moreover, in line with Miller's proposal of an \textit{evaluative AI} framework \cite{miller2023explainable}, we acknowledge the value of a machine-in-the-loop approach that privileges abductive reasoning and user reflection over mere recommendation. We concur that explanations should stimulate hypothesis formation rather than provide definitive answers. Nevertheless, certain aspects could call for deeper examination. The fact is that psychology and ethics do not necessarily align, and under conditions of stress or cognitive fragility, a form of near-recommendation may prove legitimate (what, to tell the truth, the evaluative AI framework would be ready to, being sufficiently elastic). Once an AI system assumes a prompting role, questions of opacity, responsibility, and mutual learning from error might become increasingly complex. For exmple, the possible assumption that evaluative AI could remain value-neutral would appear questionable: if models are trained on biased data, we ask whether their explanations inevitably reflect such biases. More broadly, one may conclude that models themselves possess a degree of agency --- a character of interpreters whose epistemic and ethical weight should perhaps be explicitly acknowledged.

\begin{figure}
    \centering
    \includegraphics[width=0.35\linewidth]{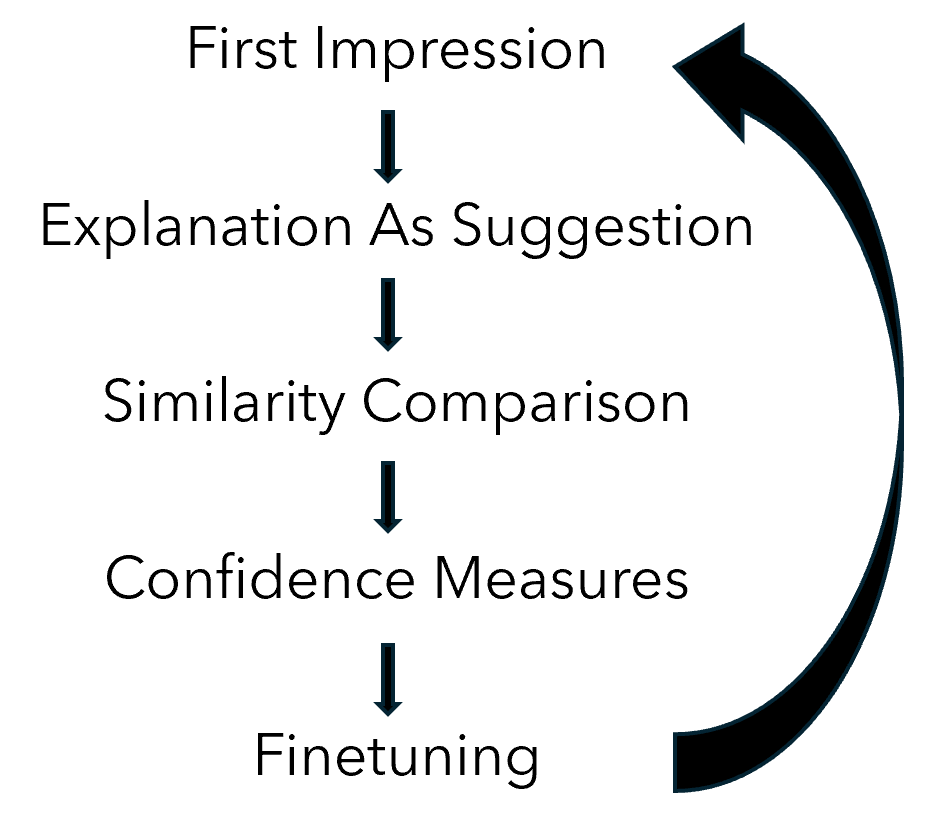}
    \caption{Diagram of the proposed Endless Tuning protocol in a decision-making setting. In order: the user is asked their first impression; is given suggestions through explanations before knowing the outcome; is given suggestions by comparing the present case with similar ones in the training set; looks at the outcome with confidences; after a sufficient number of sessions, the model is finetuned.}
    \label{fig:fig2}
\end{figure}

Our protocol consists of five components and is modular, meaning components may be individually excluded depending on specific needs. It takes the form of a dialogue with the artificial intelligent system, structured as a double negative feedback loop that can be navigated both back and forth and in a continuous flow. A multi-step interface has been designed (see Figure \ref{fig:fig2}), through which the user, guided by a series of prompts and given the option to change each time idea, arrives at a potential final decision regarding a given case (without imposing any time limitation), respectively: \textit{first impression, explanation as suggestion, similarity as suggestion, confidence measures, and finetuning}. 

Now, let us dive into the protocol, in-depth analyzing each step. 

\subsection{\textit{What's your first impression?}}  
The interface presents the user with the choice of a case --- such as an image, a row from a .csv file, and so on. Once a case study is selected, if the user feels confident about the appropriate label, they may choose to skip directly to step 5. Otherwise, at their own pace, they are first invited to form an initial impression by observing the features of the case. They should also be given the opportunity to express their thoughts through written input, such as annotations, which serve both as a means of reflection and as a potential tracking activity. \textit{This holds for steps from 1 to 4}.

Requesting the user’s initial impression is a crucial step, as it grants them the agency to begin engaging with the task. Two nuances of this concept are worth emphasizing. First, calling the user into action serves as a nudge toward cognitive effort and the assumption of responsibility. One might reasonably object that a disengaged user could simply select a random label --- perhaps even going so far as to deliberately wait a set amount of time in an attempt to game the system. Indeed, tracking devices are needed capable of capturing what actually happens without any prejudice: the user will then have to justify their actions in a credible and transparent manner. 

A second nuance concerns the very notion of \textquotedblleft beginning''. This first step is something of a false start --- or better, an aporetic beginning \cite[pp. 73-76]{fabris2016relazione}: at minimum the third, fourth, or fifth such beginning, following the initial data collection, its labeling, the development of the model, and subsequently the model's automated training. Such \textquotedblleft aporia of the beginning'' already signals the reality that our relationship with an artificial intelligence system is, in many ways, endless --- both vertically, as it may unfold in an infinite loop, and horizontally, given the length and complexity of the supply chain embedded within the product we are handling.

\subsection{\textit{I'll give you a suggestion: explanations, with a hermeneutic approach}}
\label{sec:3.2}

As a premise, in \cite{fabris2024towards} this step was originally placed third, following the similarity comparison. However, the earliest trials of the interface, informally conducted, revealed that users, perhaps due to some \textquotedblleft cognitive laziness of their System 2'' \cite{kahneman2011thinking}, tended to skip the similarity cues altogether, instead hastily placing significant weight on the allure of saliency maps. As a consequence, this sequence was experimentally inverted, which proved to be more effective. 

At this stage, the model has already produced an outcome, but this second step involves presenting the user either with an account of the process that allowed the model to reach its conclusion (\textit{e.g.}, rule-based explanations for decision trees) or with indicators of potential significance w.r.t. input data (\textit{i.e.}, saliency maps), but \textit{before knowing the outcome of the model}, avoiding \textquotedblleft apophantic'' interfaces, due to some reasons. Firstly, \textit{in the wild}, it cannot be taken for granted that the user knows the ground truth. Secondly, an explanation is a double jump: it is not relative to the input data, but to how the machine saw it. On one side, there is no absolute guarantee that the model’s prediction is correct; on the other side, the explanation itself might be unfaithful. Moreover, if the model \textquotedblleft reasoned'' well (and explanations might be managed, see f.e. \cite{teso2019explanatory}) it can provide useful suggestions; whereas, should it have reasoned badly, the user should not yet be so affected by automation bias and nonetheless be able to find out some interesting patterns, some \textit{anomalies}, as telling them: \guillemotleft hey, look \textit{here} and \textit{there}\guillemotright.  
 
 A \textit{hermeneutics of artificial intelligence} is carried out \textit{in nuce}, using explanations as interpretations. However, are such a term and such a choice justifiable? After all, once the desire to avoid the replacement of the user by the machine has been established, one might argue that there is no need to go so far. After all, the local rules extracted from a decision tree are faithful \textit{by design} and quantitative metrics such as stability and fidelity exist to assess black boxes' explanations \cite{bodria2023benchmarking}. At first glance, then, there does not appear to be that necessary distance between the signifier and the signified which typically characterizes such an approach (see f.e. \cite{bianco2007introduzione} for a survey). Nevertheless, approaching so-called black-boxes, the very notion of a \textquotedblleft local explanation\textquotedblright{} appears somewhat oxymoronic. An explanation, indeed, often bears the character of a strong hypothesis and thus, if not itself a general causal nexus, is at least derived from a general assumption. Instead, for example, the subsymbolic and connectionist functioning of neural networks is akin to what Aristotle referred to as \textit{symbebekos}, the \textquotedblleft concomitant'', which cannot be elevated to \textit{logos}, even though, in this case, it responds to a precise function. Nonetheless, to explain means to explain \textit{for us}.
 
On the one hand, Heidegger \cite[\textsection 32]{heidegger1962heidegger} already pointed out how our interpretation, including scientific knowledge, always operates within a pre-comprehension, a horizon of meaning, even constituting a vicious circle. Such pre-comprehension cannot directly grasp the inner workings of the algorithmic process and, furthermore, could be strongly influenced by a series of variables that escape the algorithm’s processing. On the other hand, might the XAI explainers be qualified as \textit{second-order interpreters}? A sign, Charles Peirce wrote \cite[2.228]{peirce1931collected},
\begin{center}
\small
\begin{quote}
    \guillemotleft is something which stands to somebody for something in some respect of capacity. It addresses somebody, that is,  creates in the mind of that person an equivalent sign, or perhaps a more developed sign. That sign which it creates I call the \textit{interpretant} of the first sign. The sign stands for something, its \textit{object}, […] not in all respects […] but in reference to a sort of idea, […] the \textit{ground}\guillemotright.
\end{quote}
\end{center}

Now, as an example, let us take the INTGRAD algorithm \cite{sundararajan2017axiomatic}. It generates a local explanation by progressively sending a spectrum from a baseline to the chosen input (say, the sign) to a black-box (global function: first piece of the ground), calculating the path integral (chosen process: second piece of the ground) with respect to the features, and thereby extracting their significance. Since the space \( x - x'\) only represents a small portion of the entire possible input space, it seems that the explanation will inevitably be affected by a bit of \textit{vagueness}. Still, to further expand such point, the fact remains that multiple valid explanations are possible (see \cite{breiman2001statistical} on the \textquotedblleft multiplicity of good models''), and we are not only referring to forms of instability caused by interpretation attacks (that is, when two slightly different input instances produce very different explanations). Rather, even when equally good metrics are achieved, if (as will be shown in Section \ref{sec:chap5}) an explainer like RISE \cite{petsiuk2018rise} observes the behavior of an entire neural network, while Grad-CAM \cite{selvaraju2017grad} observes only a single layer thereof, why should we expect them to provide the same explanation? In this play of \textquotedblleft one, none, and a hundred thousand'' black boxes, it seems appropriate to explicitly highlight some distance between signifier and signified. Not to mention the field of textual explanations, to which hermeneutics would be naturally suited, the very concept of a saliency map is intriguing. Its \textquotedblleft filagree'' (or perhaps \textquotedblleft watermark'' might be a better term) is bounded both by computational constraints (\textit{e.g.} the number of masks for RISE) and by the peculiarities of our vision, which is sensitive to color and, above all, low-dimensional. In his \textit{Metaphysical Meditations}, Descartes would wrote:

\begin{center}
\small
\begin{quote} \guillemotleft When I imagine a triangle, I do not conceive it only as a figure comprehended by three lines, but I also apprehend these three lines as present by the power and inward vision of my mind,
24 and this is what I call imagining. But if I desire to think of a chiliagon [...], I certainly conceive truly that it is a figure composed of a thousand sides [...]; but I cannot in any way imagine the thousand sides of a chiliagon, nor do I, so to speak, regard them as present [with the eyes of my mind]\guillemotright \cite[p. 25]{descartes_meditations}.
\end{quote}
\end{center}

Somehow, this reminds us of that famous question: how long is the coast of Britain? Indeed, although the inner dimensionality of models is not infinite, even our intellectual comprehension reaches its limits in front of millions of parameters. The further the point of observation, the brighter and more lively the meaning appears to be captured, but as soon as we get closer, not only does our vision disappear, but even the previously vivid sense of data processing becomes more and more obscure through a labyrinthine network. 

\subsection{\textit{I'll give you a suggestion: similarity}}

Drawing from Peirce's concept of \textit{thirdness}, \cite{marcos2011sem} describes similarity in both physical and epistemological terms as a halfway relationship --- as if positioned along a branch of a hyperbola --- between identity and difference. So as to facilitate a more or less indirect dialogue between the user and the experts who compiled the training dataset, the possibility is provided of comparing the current case study with the most similar ones in the original dataset, using the original labels. On the one hand, mapping the training data space tends to shift back in time the center of responsibility, particularly if not placing conditions on the labels. On the other hand, putting a bit more power back in the algorithm's hands, clustering algorithms might be applied, for example on a latent space. Yet, as experience taught us, \textit{it is a matter of subtle and tricky balance}.

An objection may arise, already raised by \cite{marcos2011sem}: can similarity be computed? After all, it has a formal nature, whereas we cannot help deeming it a geometric proximity measure. However, not only does the activity of a subject (the data annotator) not seem absent at all, but nothing prevents f.e. an autoencoder computing embeddings to reserve surprises in its latent space. In fact, to us, this last one seemed also the best way to account for difference, because it can introduce a margin of variance. Actually, differences usually emerge through criticisms (the more distant cases from the same class) and counterfactuals (the more similar cases from a different class). Such features are useful in the context of explainability, and at first glance, it might seem that the user's critical thinking could benefit from them. However, here we do not have to \textit{explain} the user's choice as the one of a black box, but rather \textit{guide} them through their decisions. The more, using criticisms or counterfactuals, which label should be taken as the basis? Since we cannot use the label provided by the classifier model (because it is still \textquotedblleft secret''), we would need to use the last label provided by the user. Such a choice would entail a risk. If we told the user, \textit{e.g.}, something like: \guillemotleft look at this other case now: it lies quite far from the one you are observing, yet it has the same label\guillemotright, if the their opinion were correct, they would be misled.

\subsection{\textit{Do you really want to know my opinion?}}

A model never explicitly says \textquotedblleft yes'' or \textquotedblleft no''; rather, it provides an indication. A typical case is regression. Nonetheless, particularly in classification settings, this indication might be structured, often through a softmax function applied to the last layer of a neural network, as a probability measure for the various options, thus functioning as a measure of \textit{confidence}. We could broadly define confidence as the model's certainty \textquotedblleft in itself'' concerning its own prediction, in contrast to its actual output (if it were solely about the latter, we might perhaps more appropriately refer to it as accuracy). This module, the penultimate of the protocol and the last of the interface, consists of presenting the user with the actual outcome of the model, which has been hidden up to this point, preferably together with its confidence. The purpose of providing the result alongside the confidence, as noted by the aforementioned \cite{zhang2020effect}, is to assist the user in calibrating their trust in the device (after all, this is a strategy both against automation bias and algorithmic aversion), while, in full alignment with the relational ethical approach proposed here, the human is required to contribute their own knowledge. However, caution is needed with overconfident systems (see f.e. \cite{zhang2023survey} for relative issues). On the one hand, therefore, it would be useful to warn the user; on the other hand, paraphrasing \cite{miller2023explainable}, isn't the soft voice of Prudence more ear-friendly than the one of Bluster?

\subsection{\textit{Finetuning}}

The user's decision then becomes the label for that case study (for example, an image), which is saved in memory ----- either anonymously or not, depending on the concrete requirements. Drawing from both some temporary data and the labeled case studies, a fine-tuning dataset is gradually built. Once this dataset reached an appropriate size, it is used to fine-tune the model. Naturally, should the design phase (in case of concrete application) include the decision to entrust the initiation of the fine-tuning process to an automated system rather than a human \textquotedblleft maintainer'', it would be essential to equip the system with safety guardrails (e.g., to prevent overfitting). As previously mentioned, the rehearsal techniques are the most simple, but not the only ones --- and much space is left for Continual Learning to evolve. It is obviously up to the developer to decide how to manage data, learning procedures and so on. 

In addition, it is necessary to underline at this point a socio-technical limitation in this protocol (which is in fact a limitation of artificial intelligence itself): the need for clean and robust data. Due to such a limitation, for example, it might happen that a single user will not be able to train and fine-tune a model through such a protocol by themselves alone. Too many work sessions could be needed, or a personal computer could not have the minimum computational power available, implying the need for a remote machine which in turn would entail more privacy issues. Almost constituting an \textquotedblleft environmental'' expansion of the ethical approach pursued here, we would like to place our hope (precisely due to the lightness of the continuous learning approach) in the creation of \textquotedblleft socio-technical clusters'' --- for example, asking a team of doctors from the same hospital (or at most, doctors belonging to the same geographical area of healthcare) to join their efforts in a collaborative perspective as encouraged by the European Data Act \cite{EU2023DataAct}, seeking to benefit from locally hosted servers and getting engaged in the robustification of the data.

\section{Case studies}
\label{sec:chap4}

Three simple prototypes of Endless Tuning were developed\footnote{A Jupyter notebook of the \textit{Endless Tuning} for running the same experiments, with in addition other two lighter versions of the art style recognition and the pneumonia diagnosis tasks employing 240x240 px images, is available at \url{https://colab.research.google.com/drive/1m1mDWTlE5egzT_oSR18hHLlcwwrX401N}}, each addressing a different decision-making scenario: the approval or rejection of bank loans; the identification of the artistic style of a given artwork; the presence or absence of pneumonia in chest x-rays. Each one was tested in collaboration with a corresponding domain expert, who remained anonymous, received no reward, and nicely dedicated some of their time despite job duties. Such application domains were chosen as a tribute to a classic tripartition --- science, aesthetics and ethics --- representing three distinct approaches to ground truth, respectively: a clearly identifiable fact (diagnosis); a more elusive and ambiguous one (artwork); a delicate balance between utility, empathy, and justice, although grounded in economic considerations (loan). 

Such interviews have an illustrative value and follow the style of friendly conversations, focusing on personal experience. Indeed, since it is the user who takes the decision, they are not replaced, but it is less obvious whether they \textit{feel} replaced. Our design is, say, \textquotedblleft locally oriented''. Moreover, a statistically reliable sample is hard to obtain (in front of a potential application landscape involving users with diverse cultures, biases, and competences, running a test with 100 or 1000 individuals might not yield sufficiently meaningful results). Such a choice might lack scientific validity, but our work concerns the ethics of artificial intelligence, hardly measurable. Since the experts did not have strong competence in artificial intelligence, we maintain that simply letting them do a \textquotedblleft test run'' really wouldn’t have made much of a difference. Afterwards, that's the concrete reality of the employment of artificial intelligence.  So, the presence of the interviewer appears justified.

Our interviews followed the same structure: meeting the domain expert in person; introducing them to the core concept; letting them go through a working session up to a final decision (which could also include abstention); at the end of the session, asking them about the usefulness of the suggestions (Steps 2, 3, and 4 of the protocol), about their perception of being potentially replaced by the machine, and of assigning an overall score to the project on a scale from 0 to 5. All sessions were conducted on a laptop, and lasted between 20 and 40 minutes. The relevant information was saved, and the dialogue between the interviewer and the interviewee was recorded and transcribed (see \ref{AppendixA},\ref{AppendixB},\ref{AppendixC}).  Due to time availability, it was only possible to test a single session ----- certainly not enough to constitute a fine-tuning dataset. The fine-tuning procedure was nevertheless carried out, but solely for clarifying purposes, in order to help the interviewee better evaluate the system. Since all three versions share common details, in order not to be repetitive, we will respectively present all of them in subsection \ref{sec:chap5.1}, while assessments will be gathered in subsection \ref{sec:chap5.2}

\subsection{Prototypes}
\label{sec:chap5.1}

The first version concerns the task of granting or denying a loan to a fictitious applicant, according to some indices. As a consequence, it is a binary classification task. We used the Loan Approval Classification Dataset \cite{tai_loan_dataset} made available on Kaggle, inspired by \cite{credit_risk_dataset} and \cite{financial_risk_dataset}, and licensed under the Apache 2.0 License. The dataset consisted of a .csv file containing 45,000 applicants and 13 attributes: \textit{age}, \textit{gender}, \textit{education}, \textit{income}, \textit{employment experience}, \textit{home ownership}, \textit{loan amount}, \textit{loan intent}, \textit{loan interest rate}, \textit{loan percent income}, \textit{credit history length}, \textit{credit score}, and \textit{previous loan defaults} (plus the target: grant or deny). To obtain a perfectly balanced dataset, exceeding items were randomly removed, obtaining a training dataset of 18,000 rows. At the same time, a balanced \textquotedblleft case study set'' of 200 items was extracted, together with a \textquotedblleft temporary set'' of 1795 items used as a memory, so to say a \textquotedblleft piggy bank''. The 200 case studies were available to be chosen by the expert during the test. A decision tree was trained on the training set, reaching an accuracy of $\approx 83\% $. Since our aim was simply to provide a reasonably accurate model, in all the three cases the case study set was employed as a test set, on which accuracy reached even a slightly higher percentage.

In the second case, we made use of the WikiArt dataset \cite{wikiart_dataset}, made available on Kaggle with CC0 License, and originally sourced from \url{www.wikiart.org}. The dataset initially consisted of approximately 80,000 images, spanning 27 different artistic styles. For class balance purposes, the number of styles considered in the task was reduced to seven (those with the highest number of images): \textit{art nouveau, baroque, impressionism, expressionism, symbolism, post-impressionism, romanticism}. Out of the remaining $\approx 28,000 $ images, which constituted our training set, 70 were selected as case studies, while 420 images composed the temporary set. A 34-layer ResNet \cite{he2016deep}, \href{https://docs.pytorch.org/vision/stable/models/generated/torchvision.models.resnet34#torchvision.models.ResNet34_Weights}{pretrained on ImageNet1K}, was therefore fine-tuned on $\approx 27{,}500$ images, achieving an accuracy of $\approx 65\%$. All the images were resized to 1080x1080 px, and black padding was added so as to compensate for asymmetries. 

The third task was a binary classification problem aimed at diagnosing the presence or absence of pneumonia in chest X-ray images. In this case, we employed the Chest X-Ray Images (Pneumonia) dataset \cite{chestxraydataset}, made available on Kaggle under the CC BY 4.0 license, and inspired by \cite{kermany2018identifying,kermany2018labeled}. Originally composed of $\approx 5800$ files, due to class imbalance, the dataset was randomly reduced to 2500 items as a training set, according to the labels \textit{pneumonia: yes} and \textit{pneumonia: no}. Moreover, the case study set consisted of 100 items, while the temporary set consisted of 198 items. Also here, resizing to 1080x1080 px together with black padding were carried out. Another 34-layer Resnet, identical to the first one, was finetuned on such a training set, reaching a test accuracy of $\approx 95\% $. A couple of further layers were added to both the networks: a linear layer with \texttt{output\_size=14}, and a final linear layer with \texttt{output\_size=num\_classes}. In both the second and third experiment, black padding was felt necessary because the recognition also depended on the shape of the objects depicted. It was preferred to give the right of way to the expert, making the models train on padded images, rather than to dangerously distort the data. 

Three apps were built using the \href{https://streamlit.io/}{Streamlit} library, structured as multi-page applications. After an introductory page, where the user is informed of the task and asked to click \textquotedblleft I understand\textquotedblright, the selection of the case study follows. Already at this stage, a .csv file simulating a permanent log is automatically generated. This file records the case studies, notes, decisions, and time stamps. Every page contains some illustrative text together with warnings. The selected case is displayed --- and will be displayed at each step, to avoid the user to forget --- and the first impression is given. After that, if the expert feels unconfident and looks for help, suggestions are presented.

\begin{figure}[ht]
    \centering
    \begin{subfigure}[b]{0.497\linewidth}
        \centering
        \includegraphics[width=\linewidth]{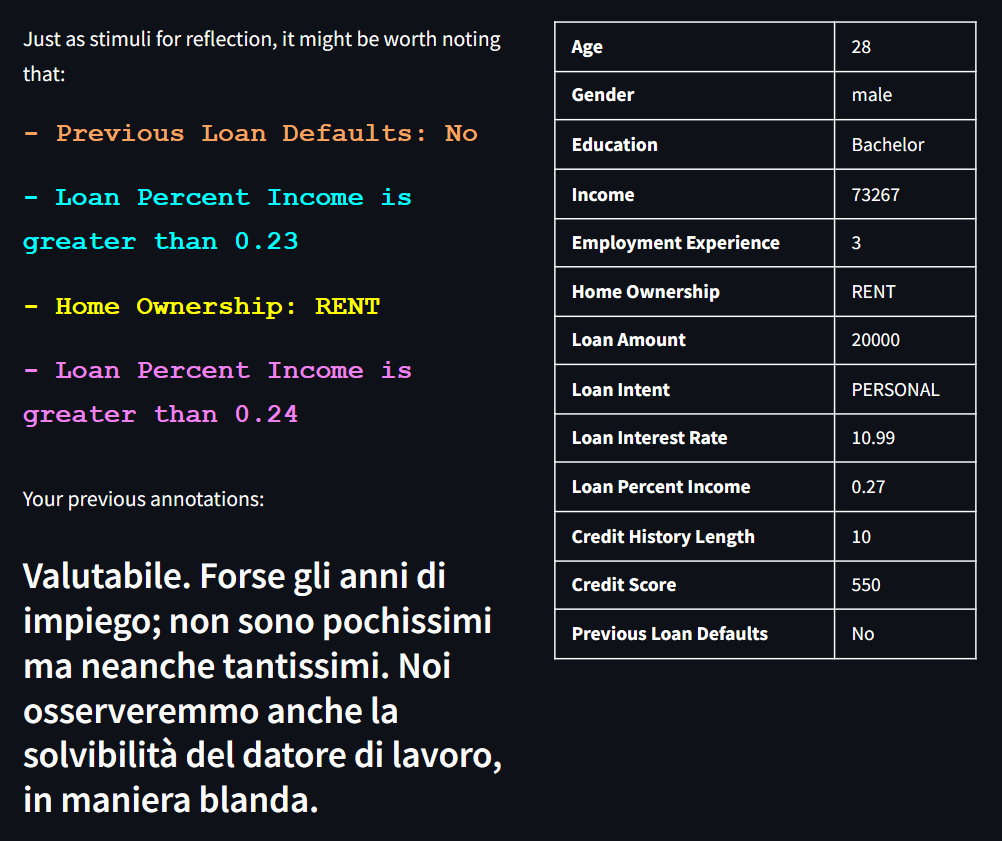}
        \caption{}
        \label{fig:3a}
    \end{subfigure}
    \hfill
    \begin{subfigure}[b]{0.45\linewidth}
        \centering
        \includegraphics[width=\linewidth]{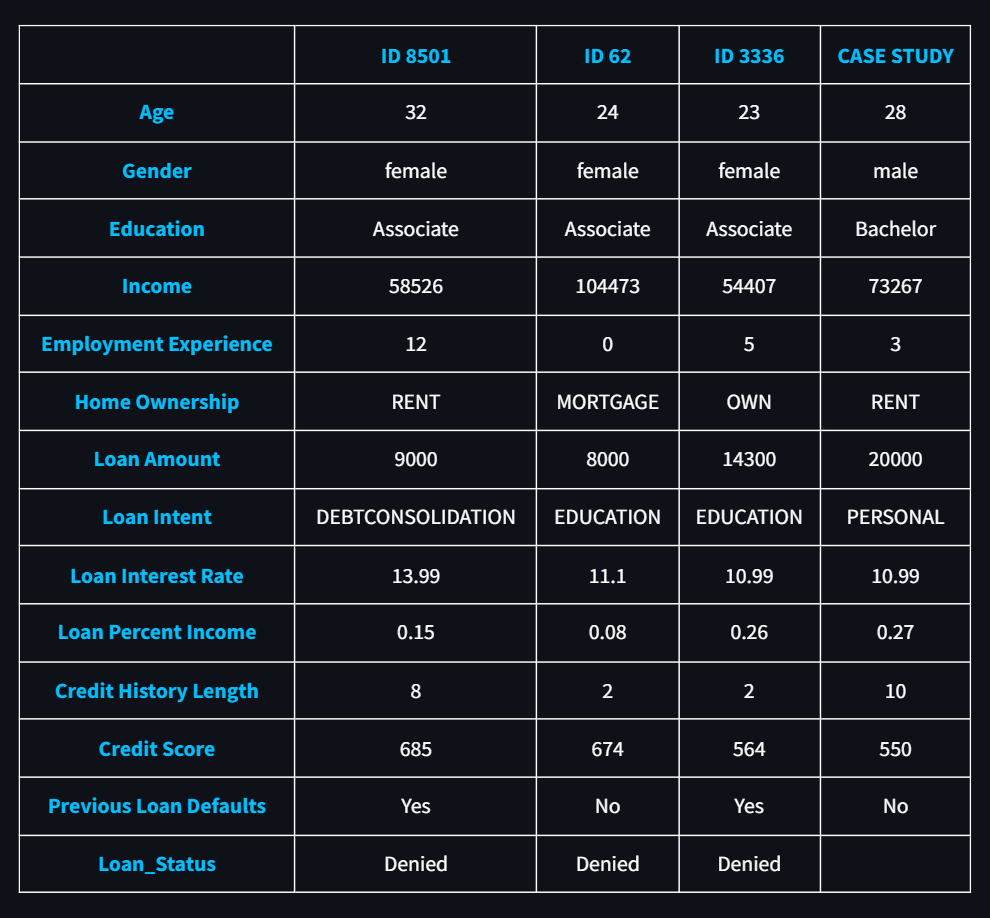}
        \caption{}
        \label{fig:3b}
    \end{subfigure}
    \caption{(a) A screenshot of the suggestion extracted from the decision tree w.r.t the case n. 144, which happened to be chosen by the bank director. The translation of the actual previous note, given w.r.t the first impression, is: \guillemotleft Worth considering. Perhaps the years of employment [in a hesitant tone]; they are not too few, but not too much either. We [the bank] would also observe the solvency of the employer, mildly\guillemotright. (b) Similarity table w.r.t case n. 144 (excluding plot).}
    \label{fig:3}
\end{figure}

In the first case, in form of decision rules extracted from the tree in a natural language fashion and variously colored --- not casually, our tree was bounded to \texttt{max\_depth=4}, so as to avoid confusion (see Figure \ref{fig:3a} for an example). Such rules are obviously truthful. Text strings were color-coded as a form of cognitive reinforcement, with careful attention to using equally vivid colors that would not be interpreted in a conventional way --- such as green for \textquotedblleft grant\textquotedblright{} or bright red for \textquotedblleft deny\textquotedblright. In the other two cases, drawing on the hermeneutic differences discussed in Section~\ref{sec:3.2}, the user is presented with two separate pages, each displaying a distinct saliency map. The first is generated using \href{https://github.com/eclique/RISE}{RISE} \cite{petsiuk2018rise} (slightly optimized due to computational constraints, producing 1,500 masks), while the second is obtained through \href{https://github.com/jacobgil/pytorch-grad-cam}{Grad-CAM} \cite{jacobgilpytorchcam,selvaraju2017grad}), applied to the GAP layer. These two visualizations reflect different perspectives, on the side of the explainers, on the neural network’s decision-making process. Both maps were carried out using the \textquotedblleft jet'' colormap, going from deep blue to intense red.

Principle Component Analysis was performed on the training data, so as to extract the embeddings and obtain similarity measures. Loan granting data had been previously scaled through a standard scaler. Instead, concerning art style recognition and medical diagnosis, PCA was applied to the vectors with \texttt{dim=14} in output from the penultimate layer of the trained Resnet-34. As for the loan granting task, a similarity table appears on the page, displaying in order the three most similar cases from the training set along with their respective outcomes (see Figure \ref{fig:3b}). As instead for the 2nd and the 3rd cases, the three most similar images according to compressed representations are plotted in the up area of the screen. Alongside them, in all the settings a Cartesian plot was shown to visualize their relative distances. 

Confidence, as probability measures extracted through \texttt{predict\_proba} and softmax, was plotted on a histogram. Finally, with regard to fine-tuning, it must be noted that, in the case of rigid structures such as decision trees, the term is perhaps somewhat inappropriate. More than in other cases, the flexibility of retraining (here forcefully understood as full retraining) depends primarily on data management. In general, the following data augmentation strategy was adopted. Whenever the user makes a decision, the corresponding case is stored in memory. At that point, another case is randomly selected from the temporary set, provided that it carries a label different from the one assigned by the user. These cases constitute the starting point for the progressive creation, session by session, of a \textquotedblleft fine-tuning set\textquotedblright. Naturally, more sophisticated alternatives to such a toy approach can be considered. Once this fine-tuning set reaches a sufficient size, a rehearsal can be performed.

\subsection{Insights from user experience}
\label{sec:chap5.2}

The first interviewee was a bank director, aged 55, with 30 years of experience in banking, 24 of which as a director. The subject appeared to be an exemplary case of a highly serious and experienced individual who, nevertheless, is not yet particularly familiar with artificial intelligence. This, contrary to what one might expect, only makes the setting more realistic. When asked whether he felt replaced by the machine, he responded, with a hint of disappointment: \guillemotleft the decision-making part, honestly, felt more mine than the machine’s\guillemotright, revealing the most common expectation---a form of \textit{automation bias} --- while paradoxically confirming that human replacement had been avoided. On the other hand, when presented with the counterfactual question regarding a possible disagreement between his own judgment and that of the decision tree, he replied: \guillemotleft I'd go back through the process and try to understand what part diverged from my assessment\guillemotright. Drawing on his professional experience, the director focused primarily on the system’s performance rather than its interface. Indeed, while he expressed satisfaction with the alignment between his own intention and that of the decision tree, he explicitly pointed out the absence of certain specific parameters, such as the amount of rent. Note, indeed, that exactly because of such flaws he responsibly decided to abstain from the final decision. This led him to assign a rating of 4 out of 5, nonetheless acknowledging the system’s reliability and appreciating the fact that a record of the operations is preserved.

With regard to the evaluation of the interface, we observed some intriguing discrepancy between the interviewee’s responses and his actual behavior. When asked about the usefulness of the rule-based suggestions (see Figure \ref{fig:3a}), the interviewee acknowledged their relevance, particularly in relation to a comparison between the \textquotedblleft loan percent income'' threshold derived from the decision tree and that applied by the bank. However, it is worth noting that this aspect became a focal point of the discussion because, while reading the suggestions, the interviewee misread a value and was corrected by the interviewer (\guillemotleft Oh right --- our bank’s threshold is 33–35\%, that’s why I got confused\guillemotright). On the one hand, this may suggest the presence of an anchoring bias (see again \cite{kahneman2011thinking}), which could indeed be mitigated; on the other hand, it reveals the need for a more distinct and prominent interface than the brightly colored text currently employed. Concerning the similarity, the interviewee answered skeptically (\guillemotleft because I evaluate each case individually\guillemotright). While it must be noted that --- both in this and in the other two cases --- the chart proved to be of little use (perhaps, we may hypothesize, because it displayed relative distances without providing a fixed point of reference), it is nonetheless noteworthy that, despite his own statement, the director spent a considerable amount of time carefully examining the similar cases (see Figure \ref{fig:3b}). Without suggesting that the interviewee was in any way unaware, this observation seems to warrant further investigation aimed at catching user attention.

\begin{figure}
    \centering
    \begin{subfigure}[b]{0.321\linewidth}
        \centering
        \includegraphics[width=\linewidth]{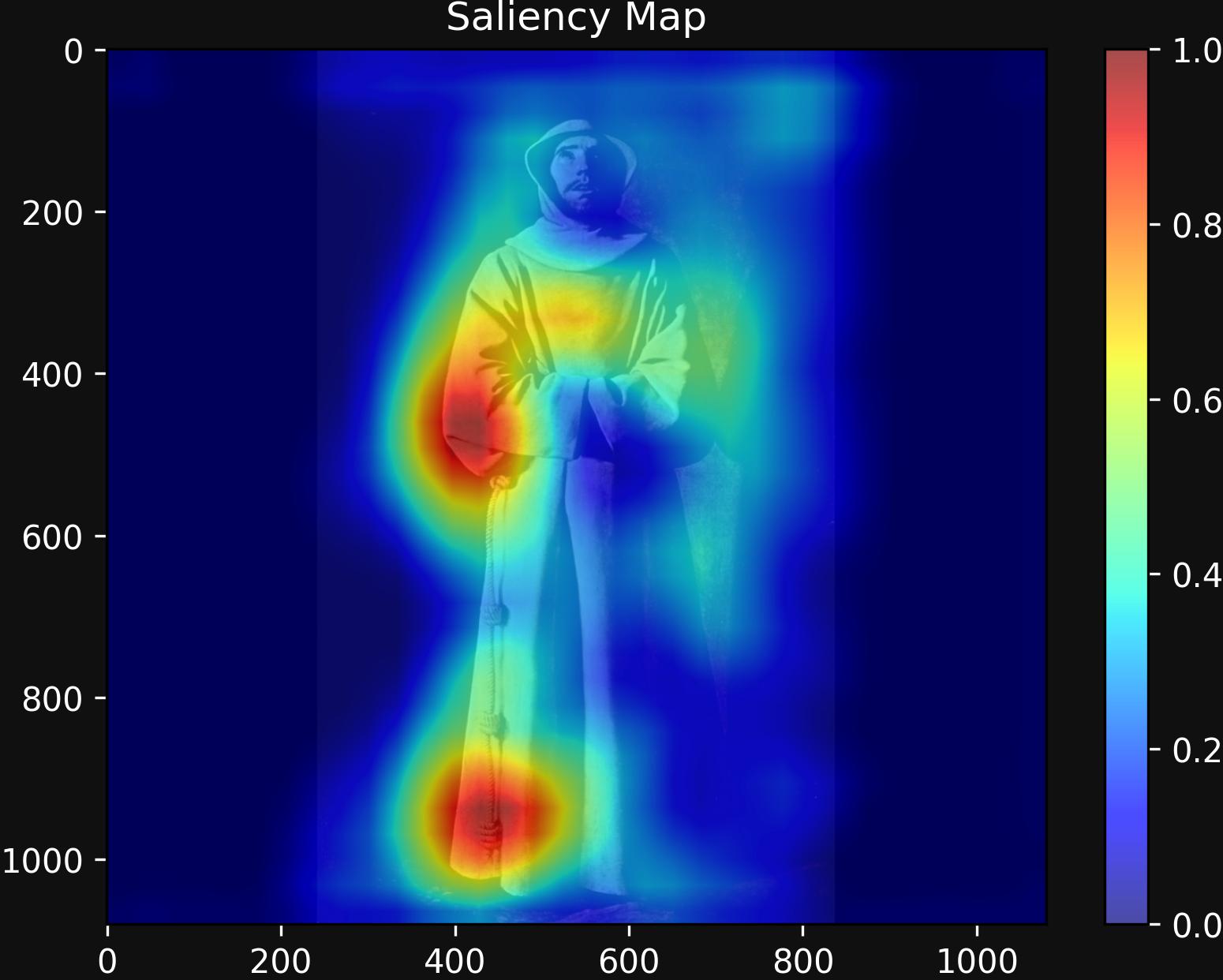}
        \caption{}
        \label{fig:4a}
    \end{subfigure}
    \hfill
    \begin{subfigure}[b]{0.33\linewidth}
        \centering
        \includegraphics[width=\linewidth]{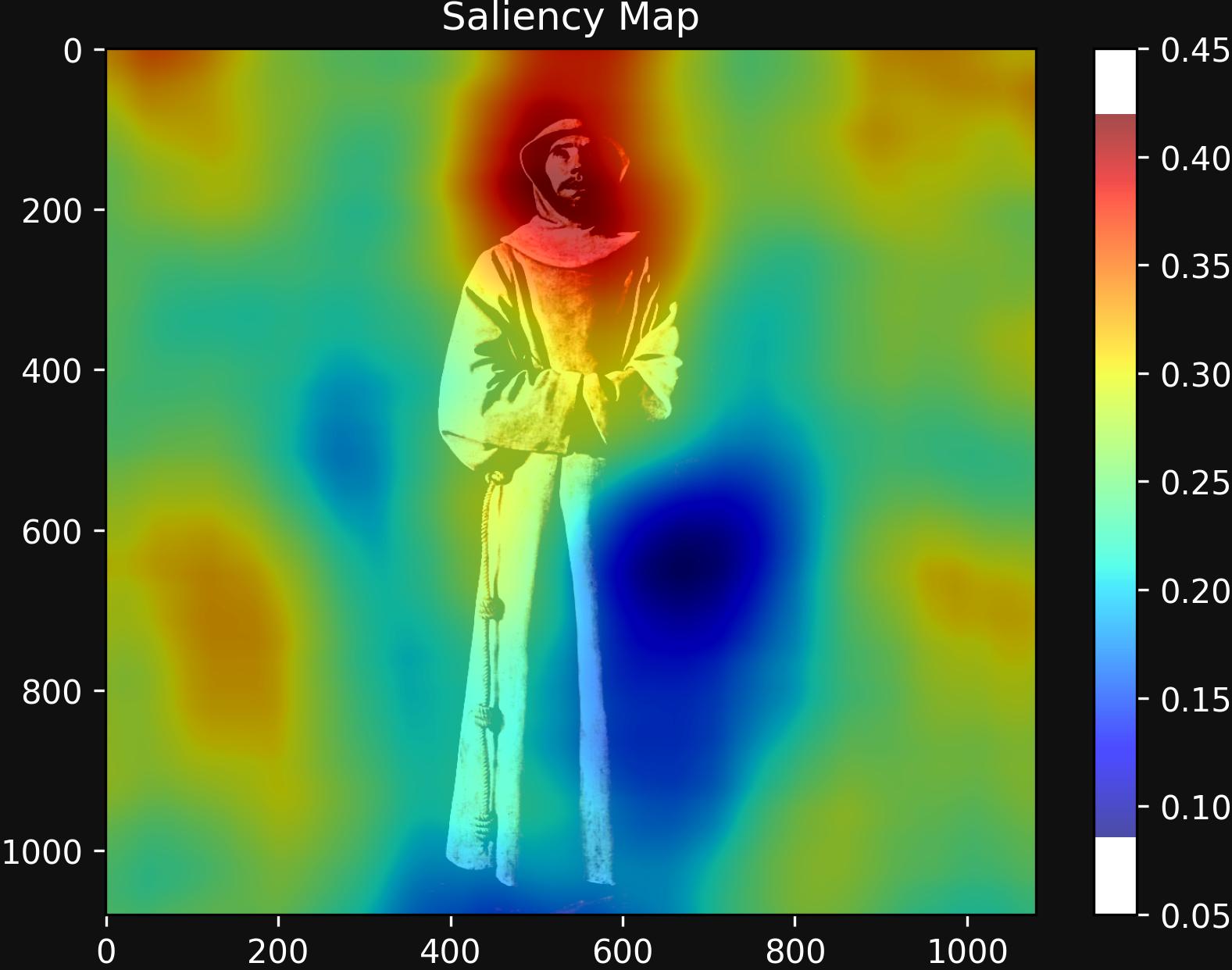}
        \caption{}
        \label{fig:4b}
    \end{subfigure}
    \hfill
    \begin{subfigure}[b]{0.26\linewidth}
        \centering
        \includegraphics[width=\linewidth]{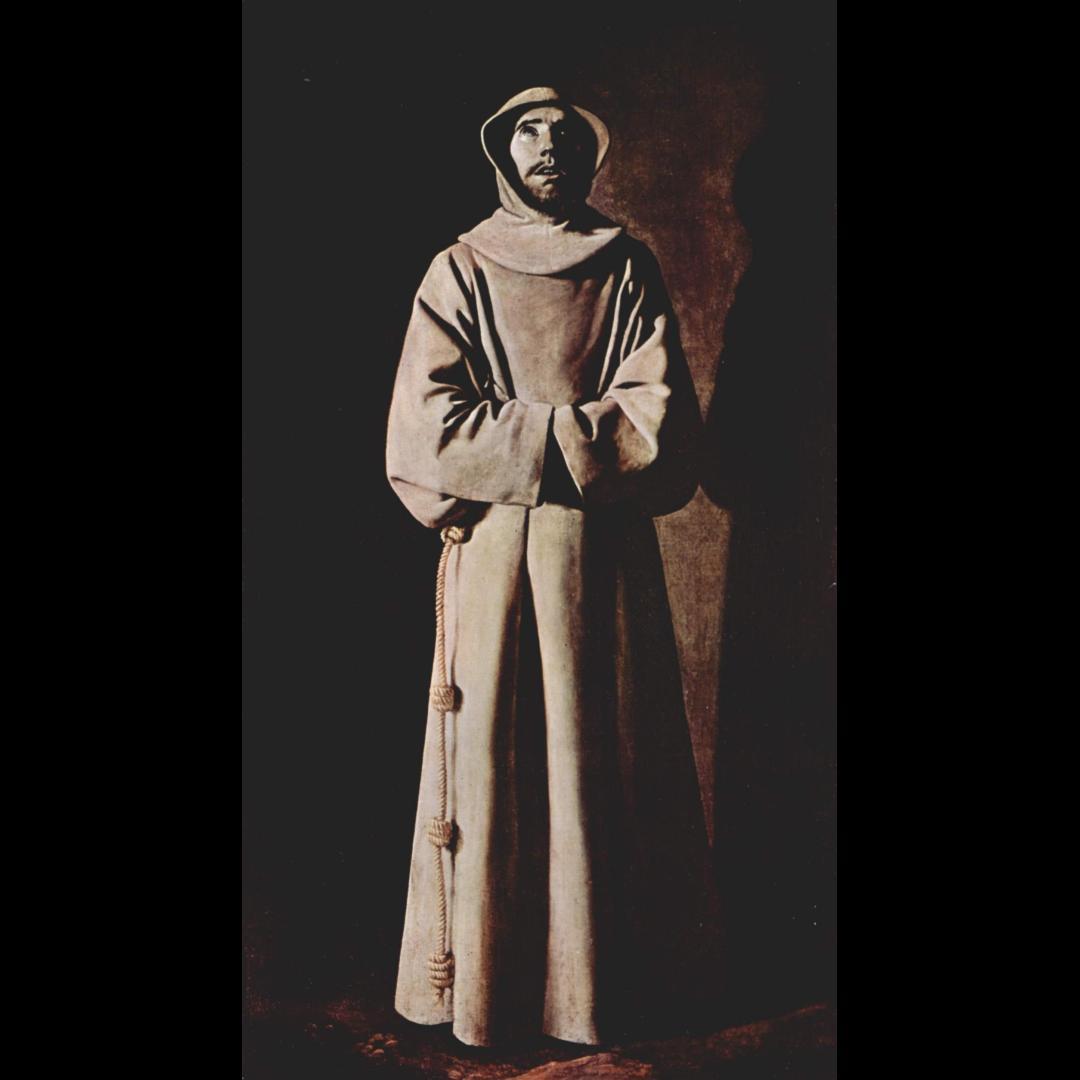}
        \caption{}
        \label{fig:4c}
    \end{subfigure}  
    \hfill
    \caption{(a) The saliency map produced with the \href{https://github.com/jacobgil/pytorch-grad-cam}{Grad-CAM} \cite{jacobgilpytorchcam,selvaraju2017grad} explainer, regarding the picture chosen in the art style recognition test. (b) The corresponding \href{https://github.com/eclique/RISE}{RISE} \cite{petsiuk2018rise} explanation (both maps presented before knowing the outcome, exploring a hermeneutic strength). (c) The picture chosen by the professor of aesthetics. Notably, it is a painting of 1659 by Francisco de Zurbarán representing Saint Francis.}
\end{figure}

\begin{figure}
    \centering
    \includegraphics[width=0.5\linewidth]{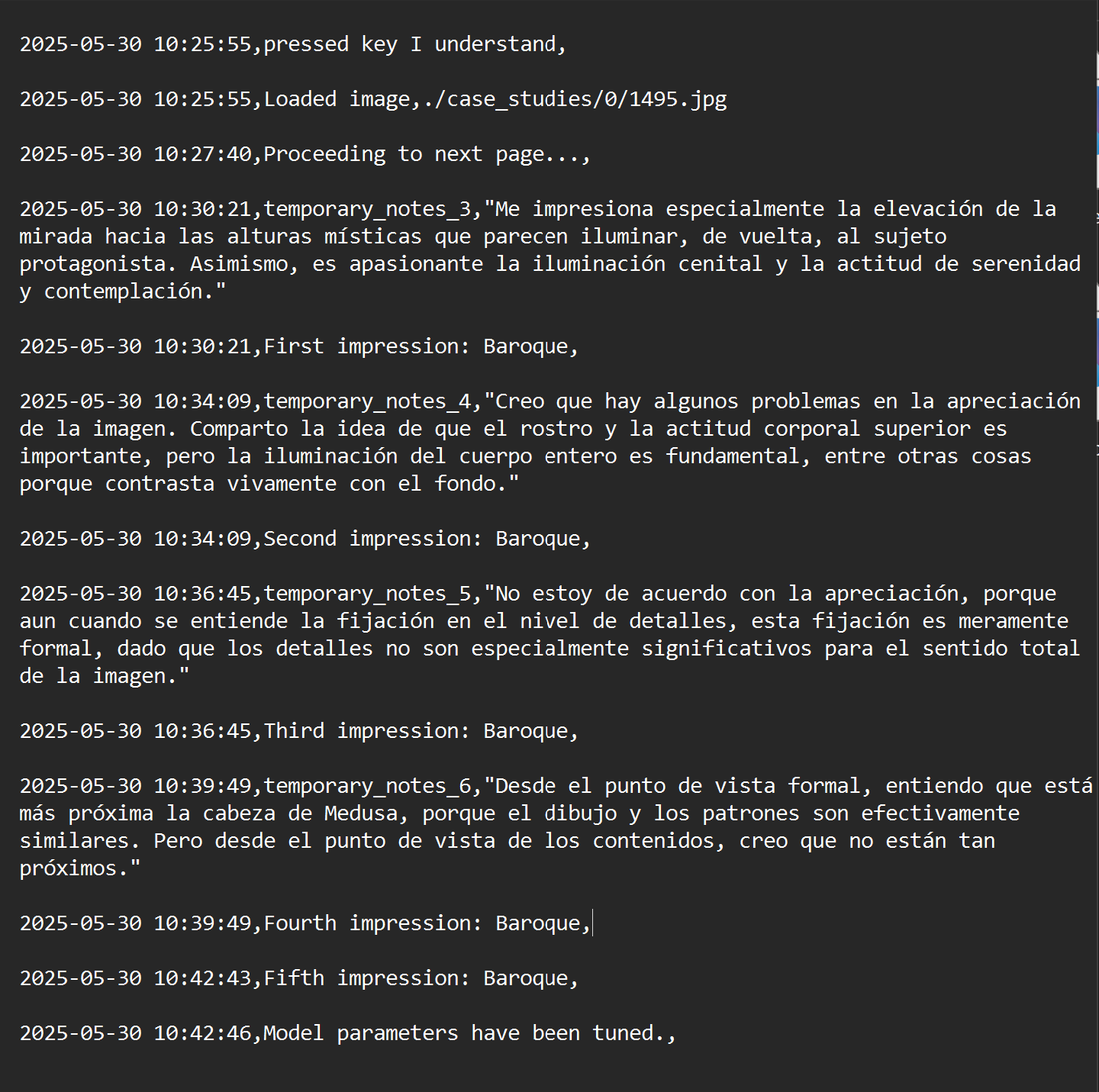}
    \caption{An example of the recorded log (art style recognition test).}
    \label{fig:log}
\end{figure}

The expert assigned to the second task is a 46-year-old professor of aesthetics with 20 years of experience. The interviewee expressed enthusiasm about participating in the experiment, intrigued by the application, and after a few minutes of getting accustomed to it, became increasingly aware of the task, realizing that they were \textquotedblleft dialoguing'' with the system. Since it was a cordial conversation, the steps were gradually commented on together with the interviewer. The professor of aesthetics contributed the most with written annotations, as allowed by our design, providing detailed guidance (see Figure \ref{fig:log}. The response provided by the artificial intelligence model eventually matched that of the expert: \textquotedblleft baroque'' class. However, the core of the experience seems to lie in the comparison with the saliency maps. Not only did the user understand the difference in approach between RISE (importance clouds) and Grad-CAM (focus on details), but they also engaged in a sort of comparison between these two explainers, confirming their hermeneutic significance. 

Indeed, regarding the first suggestion, they write: \guillemotleft I believe there are some issues with the evaluation of the image. I agree with  the idea that the face and upper body posture are important, but the lighting of the whole body is crucial,
partly because it contrasts vividly with the background\guillemotright. And, above all, concerning the second suggestion they engage in a relationship, since the expert reasoned about the picture with the pair model-explainer, yet not calling them a \textquotedblleft thou'': \guillemotleft No, no, I know the image. So... I’m not sure about this. No. Perfect. No. Because I see the sense of focus, because in fact there are a lot of details in these signs but they are not significant. I’m thinking to the whole of the image\guillemotright. When asked about the usefulness of such suggestions, they replied positively: \guillemotleft Yes. [...] Because it’s a copilot for discussing with it. Same thing when I translate from Latin, for example, and I use LLMs and I’m translating like ever, like always. And I say, okay, LLM, what do you think about this translation? [...] And I think that it can be useful for anyone that doesn’t know about... It’s not so explicit\guillemotright. 

With respect to similarity (the comparison with the three most similar images) the user expresses themselves as follows: \guillemotleft I believe that from a point of view... from a formal point of view, I think that it’s... It’s right, but according to... According to a content point of view, content, sense, meanings... [...] From the formal point of view, I understand that it’s closer to the head of Medusa, because the drawing and the patterns are actually similar. But from a material point of view, I think they’re not so close\guillemotright. When asked about an assessment, they replied that they were surprised and that they did not exactly understand why such images were chosen. It did not seem precisely that such a cognitive forcing strategy cannot work at all: clues were not deemed as insignificant. Rather, the precision of the embeddings and the accuracy of the model itself are crucial and should be carefully balanced with the expertise inherent to the training set. Such balance was not reached in our prototype, indicating a weak point. Indeed, the extraction of \textit{tiny} vectors from a latent space (what the interviewee referred to as \textquotedblleft material aspect\textquotedblright) almost ended up conflicting with the original labeling of the data, neither allowing the annotators' expertise to shine through, nor putting in evidence some sensible anomalies.  

Finally, they gave 4 out of 5 as an overall evaluation and strongly affirmed to have felt full control over the process, without being replaced.

The expert involved in the third task was a radiologist, aged 59, with 27 years of experience. Despite not having much literacy on artificial intelligence --- it was indeed the first test he had ever made with such a technology, at least in the medical field, and that is the reason why they gently declined to score the project (\guillemotleft Honestly, I wouldn’t know how to rate it, [...] and I wouldn’t know how to weigh the pros and cons\guillemotright) --- the doctor showed a sort of conscious skepticism, yet not provoked by knowledge about vulnerability in models or data. Rather, they even doubted about the so-called automation bias (\guillemotleft In my opinion, it's the opposite\guillemotright), seemingly due to self-confidence, which nonetheless appeared to be grounded in concrete meticulousness. Suffice it to know that they reviewed the picture multiple times, zooming in on details, and comparing such details with saliency maps and other suggestions. A clue to such an interpretation might be provided by the fact that they strongly posed the question of whether the responsibility for damages were to be ascribed to the physician or not, expressing some doubts towards the efficacy of a judge regarding the settlement of a lawsuit, in favor of the deployment of precise standards of trustworthiness.

\begin{figure}
    \centering
    \begin{subfigure}[b]{0.321\linewidth}
        \centering
        \includegraphics[width=\linewidth]{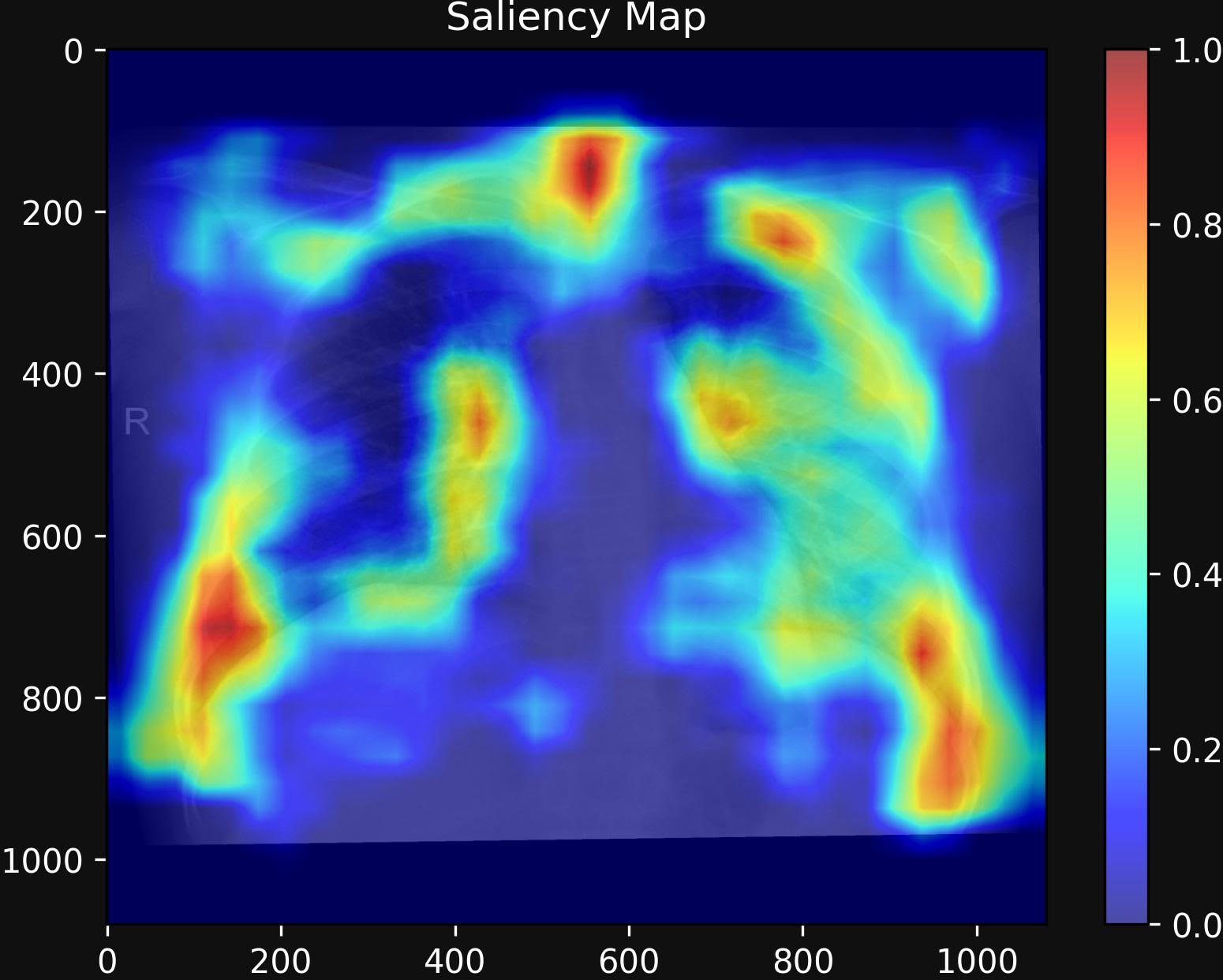}
        \caption{}
        \label{fig:5a}
    \end{subfigure} 
    \hfill
    \begin{subfigure}[b]{0.33\linewidth}
        \centering
        \includegraphics[width=\linewidth]{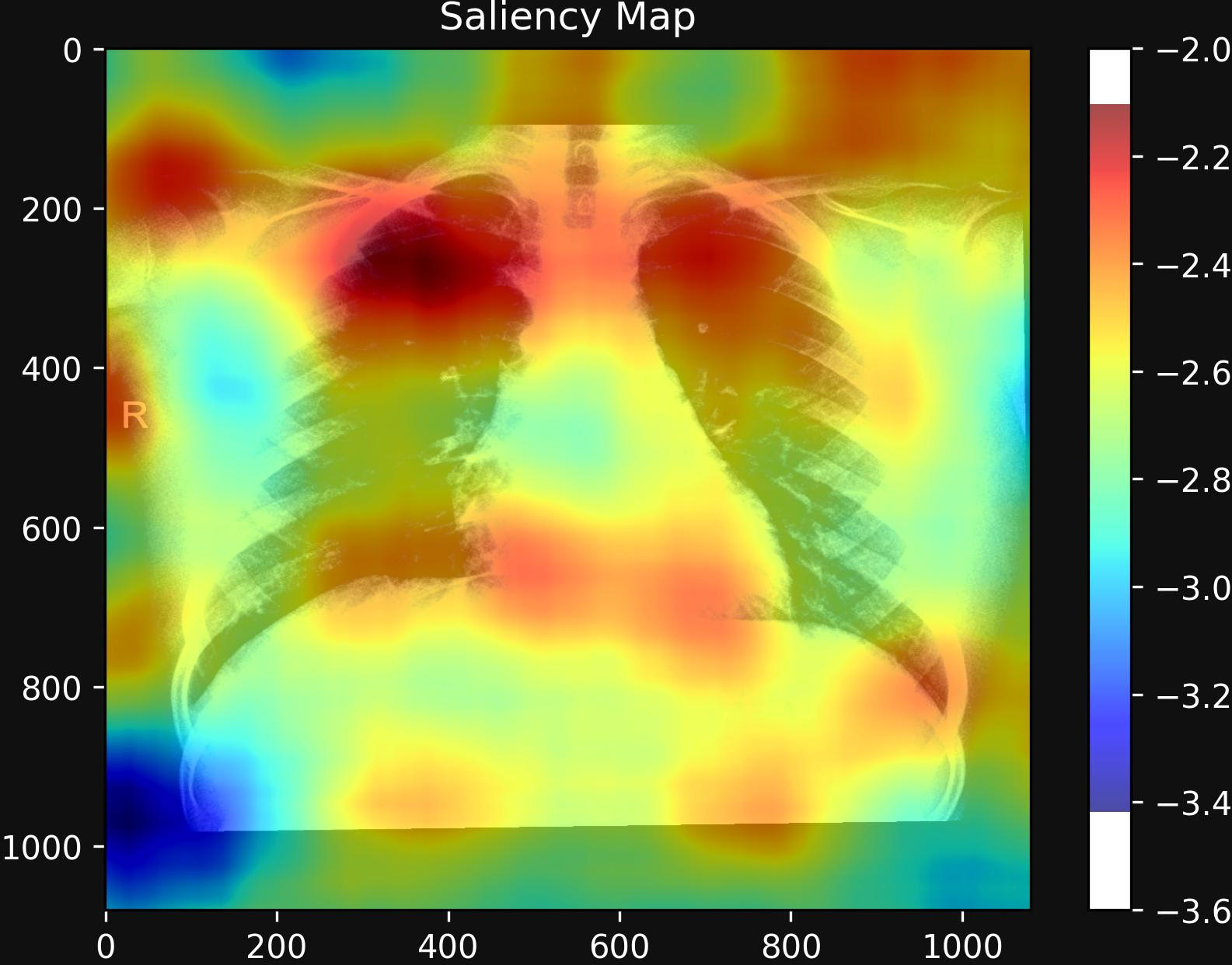}
        \caption{}
        \label{fig:5b}
    \end{subfigure}
    \hfill
    \begin{subfigure}[b]{0.26\linewidth}
        \centering
        \includegraphics[width=\linewidth]{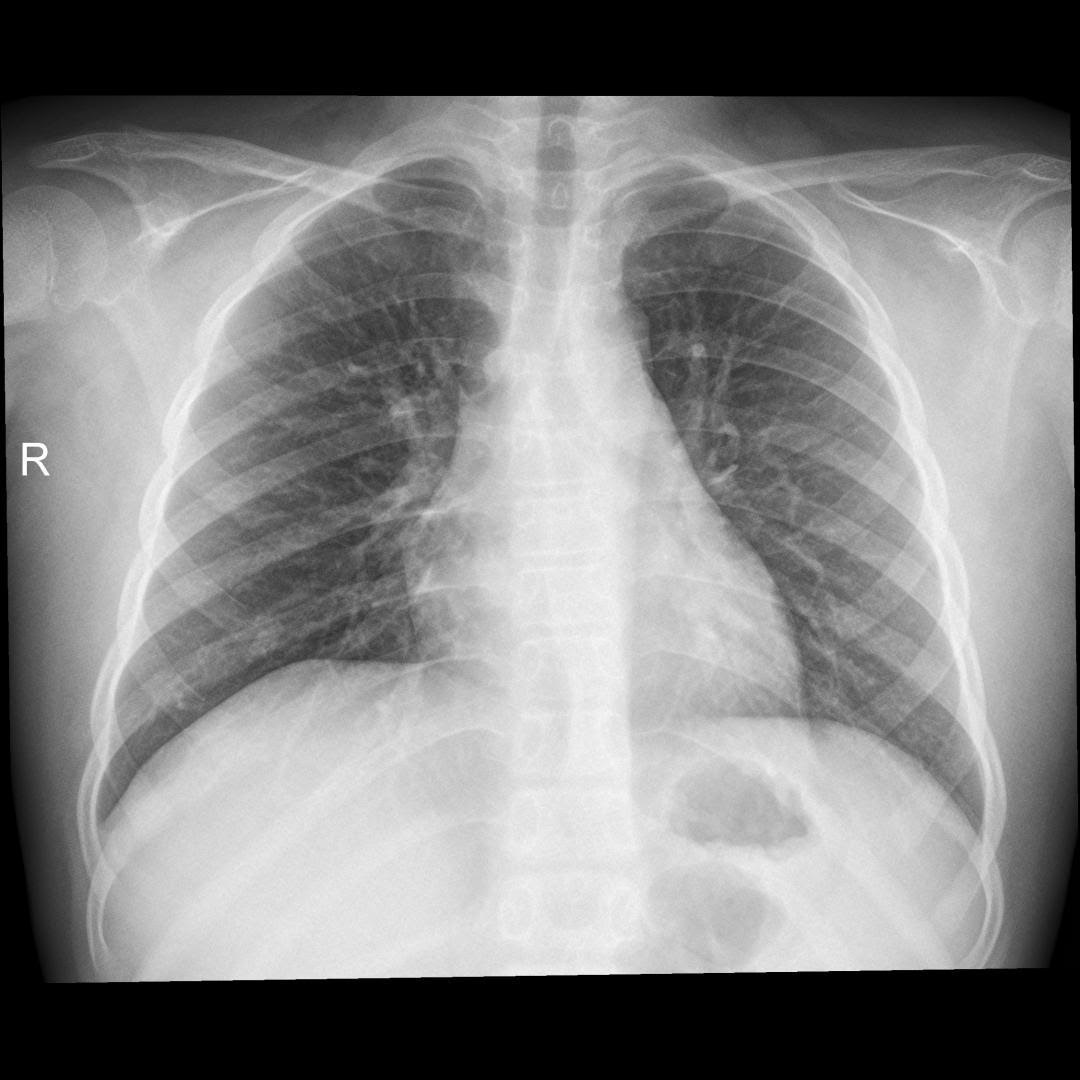}
        \caption{}
        \label{fig:5c}
    \end{subfigure}  
    \hfill
    \caption{(a) The saliency map produced with the \href{https://github.com/jacobgil/pytorch-grad-cam}{Grad-CAM} \cite{jacobgilpytorchcam,selvaraju2017grad} explainer, regarding the picture chosen in the pneumonia diagnosis test. (b) The corresponding \href{https://github.com/eclique/RISE}{RISE} \cite{petsiuk2018rise} explanation (both maps presented before knowing the outcome, exploring a hermeneutic strength). (c) The chest x-ray picture chosen by the radiologist.}
\end{figure}

\begin{figure}
    \centering
    \begin{subfigure}[b]{0.45\linewidth}
        \centering
        \includegraphics[width=\linewidth]{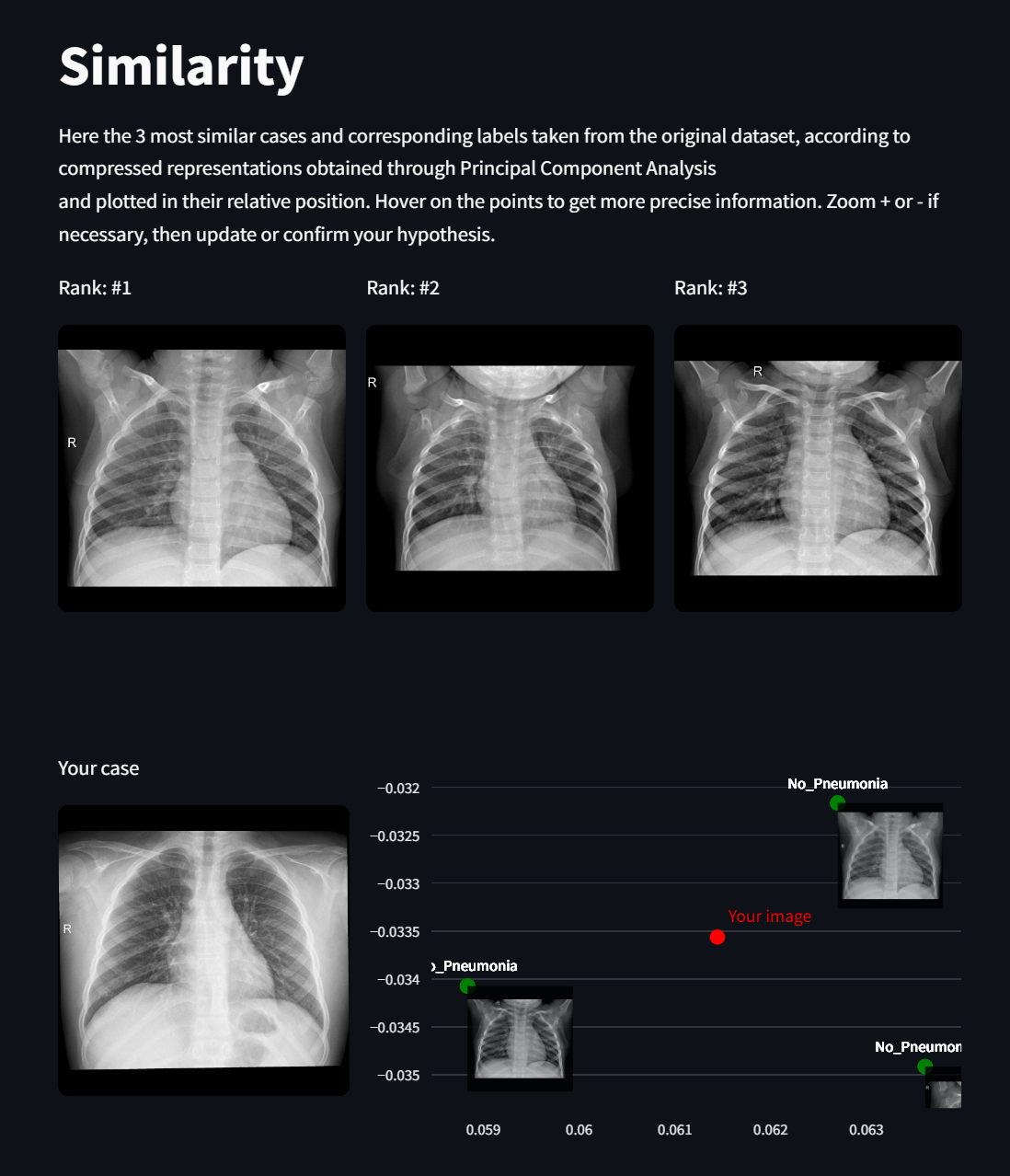}
        \caption{}
        \label{fig:6a}
    \end{subfigure}
    \hfill
    \begin{subfigure}[b]{0.5\linewidth}
        \centering
        \includegraphics[width=\linewidth]{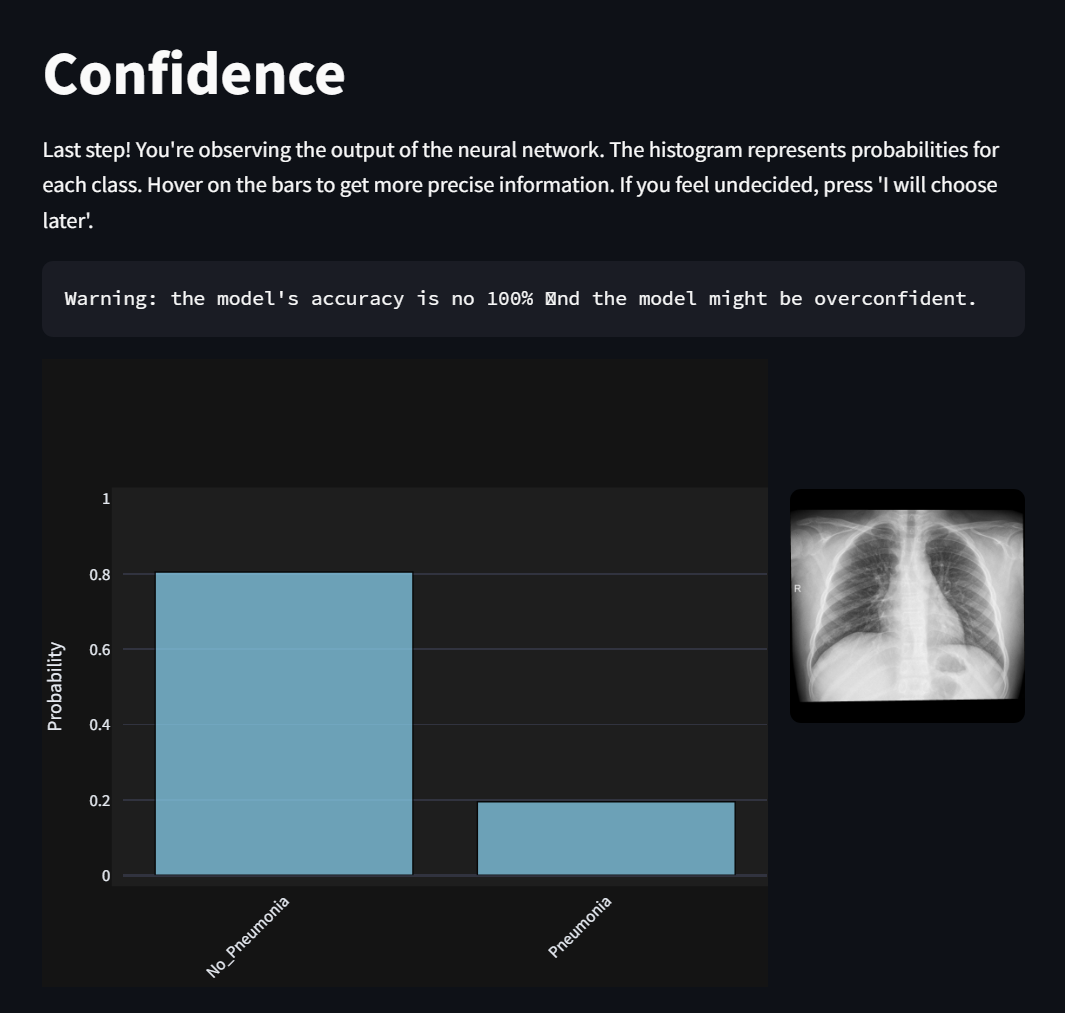}
        \caption{}
        \label{fig:6b}
    \end{subfigure}
    \hfill
    \caption{(a) The core of the similarity page w.r.t. the pneumonia diagnosis task. (b) The core of the confidence page of the same case study.}
\end{figure}

The suggestions provided by the two XAI algorithms (see Figure \ref{fig:5a}, \ref{fig:5b}, \ref{fig:5c}) put in evidence a concrete negative possibility: that is to say, that the machine makes a good prediction but \textit{for wrong reasons} (we take the liberty of saying so because we know the original label of the selected case study, which was confirmed by the interviewee and by the label attributed to similar cases). The radiologist dryly affirmed: \guillemotleft The maps were completely inaccurate: what the system flagged was completely unreasonable ---–- it highlighted features that couldn’t possibly mean anything.\guillemotright If on the one hand this may be partially due to unfaithfulness on the side of RISE and Grad-CAM (at least according to how we used them), on the other hand it put in evidence how badly we sometimes trust artificial intelligence (in our case, remember that only 2500 images were used for training). The combination of XAI and the expertise of the doctor proved to be effective in making a trust assessment.  As for similarity (see Figure \ref{fig:6a}), while the chart was deemed to be of limited usefulness, it is equally true that the step was completed swiftly, as the images identified as similar had previously been labeled with the same classification provided by the physician, resulting in full agreement on a straightforward case. Lastly, however, the prediction was reasonably correct with a confidence of 80\% (see Figure \ref{fig:6b}), and the interviewee declared not to have felt replaced by the device in any way.

Since the interviewee did not provide an explicit score, it is a bit more difficult to directly extract some final considerations. Nonetheless, it is an occasion to make a reflection upon our design in a worst-case setting --- which is and actually always will be a possibility. With respect to the saliency maps, we had poor and \textit{senseless} suggestions, yet providing the correct answer. Now, if the user (as it was our case) is expert, first of all, they can simply notice that. XAI allows the trustworthiness o\textit{f the entire system} to increase. In other words, admitting that explanations are faithful, precisely some distance between the signifier (the model) and the signified (the chest x-rays) can be correctly reported. Passing to the last step, the one which involves the actual output of the model, the user might find themselves in front of a wrong or a correct prediction. It is evident that in such a situation the user will reasonably have some reluctance in considering that output. They will have managed to calibrate their trust, either being aware that saliency maps were actually computed as \textit{post-hoc} explanations or still considering them \textit{ante-hoc}, maybe making a less precise calibration. It might well happen that the system would be excessively distrusted, in a case where the output of the model is instead correct. However, apart from the fact that abstention should be possible, given that the ground truth is unknown, even if the model was right it seems perhaps less momentous to contradict it. Indeed, it could be right \textquotedblleft by chance'', while the user might nonetheless have some reason to express, therefore governing the error.

But let us take a user who would be dragged by sensible but wrong suggestions. In such a case, there would be a real danger of being fooled. The point, however, is that when the user sees the map, they have no way of knowing what the highlighted areas are actually evidence of. Therefore, there remains a certain margin within which the deception can still be managed. If, in the end, the model turns out to be right --- albeit for the wrong reasons, but reasons nonetheless ----- and manages to persuade an especially optimistic user, so be it. On the other hand, if the model is wrong, then within that presumably narrow space for possible misguidance, the user might not bear responsibility.

\section{Filling gaps}
\label{sec:chap5}

Such a design possesses a somewhat vertical character and appears well suited to medium-to-high-stake situations. Our aim is to place emphasis on the trust granted \textit{to the user for what they are}, through affordances too, continuously re-building trust itself or perhaps, using a more concrete and engaging term, enhancing \textit{reliability}. A valid objection could be that such a relationship must be oriented toward improvement, avoiding the decay of information. Now, if an artificial intelligence system manages to provide \textit{useful} clues that are \textit{meaningful in their own way}, such information cannot be elevated to the status of knowledge. Strictly speaking, \textit{this}, to us, is the ethics of artificial intelligence. The genitive is objective, not subjective. One should contribute something of their own: a computer program alone cannot solve all our problems and yield honest and serious and competent people out of us. As previously mentioned, a slightly larger margin of error might be tolerated --- whereas, statistically speaking, it would otherwise always be deemed as a fortuitous case --- if it would be governable: the interviewees reported full control. Broadly speaking, an approach oriented to ethics of care seems to emerge, where principles certainly matter but, without underestimating the centrality of fundamental rights, it is acknowledged that \guillemotleft the morality of rights differs from the morality of responsibility in its emphasis on separation rather than connection, in its consideration of the individual rather than the relationship as primary\guillemotright \cite[p. 19]{gilligan1993different}. In short, ethics is conceived as a field with a variable center of gravity. \textit{In the beginning is the relation} \cite{Buber1970-BUBIAT}. It is primarily the operator and, more or less directly, the dataset annotators, developers, interface designers, etc. who relate to a potentially harmed party, except for the reversal of the burden of proof by the entire supply chain. As a consequence, time plays a role: responsibility is created \textquotedblleft live\textquotedblright, in the very acting, and it is a responsibility \textit{of} using the device but also \textit{for} the people (note, in passing, that the product's flaw has itself a relational nature \cite[pp. 240 ff.]{dal2024intelligenza}). On the other hand, it does not simply hold that \guillemotleft the journey prevails over the object\guillemotright{} \cite[p. 153]{virilio1989macchina}\footnote{Translation is mine.}. Such temporality might be perhaps described as a \textquotedblleft delayed virtuality\textquotedblright.

But let us focus on the design issue: namely, responsibility within the limits of the use of our tool or product. A potentially harmed party is a vulnerable subject who enters the loop insofar as the data analyzed in a session represents them, because \textquotedblleft they are being spoken about''. With regard to the legal aspect --- of course, merely attempting to extend a hand to the jurist --- as described in Section \ref{sec:chap2} and easily exemplified in \ref{sec:chap5}, the Endless Tuning envisages the presence of a permanent log, accessible only to the authority, which works like a black box on an airplane or an industrial plant, where certain valuable information is recorded so as to retrospectively reconstruct what happened. Specifically, all the information obtained in a non-fully deterministic manner should be saved. A provisional list might include: the pretraining and finetuning datasets (with documentation pertaining to them); the model weights according to the various trainings; the accuracy and loss metrics; the case studies; the feature importance outcomes (if generated through random processes); and all the user's interactions, taking into account time, date, annotations etc. The idea is that, in case of a lawsuit, a judge should be able to review the entire process in \textquotedblleft slow motion''.

Such a project seems to align with \cite{comande2018responsabilita} when, concerning a \textit{computational accountability} to trigger a \guillemotleft multilayer accountability system\guillemotright (also referring to a definition of accountability from a computational perspective by the IEEE \cite[p. 3]{ieee_glossary2017}), it contends to be possible \guillemotleft for computing itself to enable the almost automatic verification of the causal reasons behind AI actions\guillemotright[p. 1011]\footnote{Translation is mine.}, so that the \textquotedblleft bill'' could be effectively distributed. Surely, the AI Act fits this approach, providing precise obligations towards safety \cite[p. 360]{dal2024intelligenza}, but a link exists too to the other face of responsibility, \textit{i.e.} liability. As an example, the core of the AI Liability Directive proposal, actually retired by the EU Commission in \href{https://www.ansa.it/osservatorio_intelligenza_artificiale/notizie/progetto_solaris/2025/02/12/la-commissione-ue-ritira-la-direttiva-sulla-responsabilita-da-ia_00ae0807-d41d-4ed5-a225-1d938d27af4b.html}{2025}, nonetheless consisted of providing \guillemotleft an economic incentive to comply with safety rules\guillemotright, contributing \guillemotleft to the enforcement of the requirements for high-risk AI systems imposed by the AI Act, because the failure to comply with those requirements constitutes an important element triggering the alleviations of the burden of proof\guillemotright \cite{eu_com2022_ai_liability}. Nevertheless, it could perhaps be asked how straight such a link is and, again, if the structure itself of the software could not help bridging these two banks, up to touching (at the extreme side of a spectrum starting from guilt or alleged guilt) objective liability, maybe facilitating the application of the case in point of damages provoked by goods held in trust (which, at least w.r.t. the Italian Civil Code (art. 2051), seems the most appropriate \cite[p. 148]{dal2024intelligenza}).

Regarding the Endless Tuning, a design flaw (such as, for example, missing to stop the fine-tuning of an algorithm when a safety accuracy threshold has not been reached) could not only end up violating ex ante regulations, but might also become traceable. It is the protocol itself that could make it possible to link the damage not, strictly speaking, to the failure to comply with a regulation, but rather to the technical consequences that such non-compliance entails. Suppose that during the use of such a system (\textit{e.g.} for clinical purposes), an error occurs which causes harm to a person, who then decides to file a lawsuit. The authority, by retracing the log and activating all the functionalities provided --- such as reusing the model and the weights employed at the time of the relevant work session --- will be able to fully observe the process. At that point, it might for instance become apparent that: (a) the user provides vague or hasty assessments, thereby displaying a lack of skill, negligence, or imprudence; (b) the interface designer selected a colormap for the feature importance maps that tends to highlight only the most salient regions, effectively nullifying the informational value of moderately important areas of the image, in a manner that could even be misleading; (c) the developer had proposed an unreasonably small number of comparison cases in the similarity section; (d) in truth, everything suggested a prediction that ultimately turned out to be incorrect---perhaps, then, a fortuitous case. 

Having access to all this information --- does it not allow for a better understanding of who made the mistake, and where?

Naturally, such an approach presents some weak points. For example, each setting has its own timeliness: one could not state that an operator spent too little time in a session, reproaching them for their imprudence, independently of the context (\textit{e.g.} medical diagnosis \textit{vs} crop selection). Furthermore, there is the possibility that a malicious user deliberately leaves the session open merely in order to give the impression of carefully studying the case. In this regard, again, careful consideration should be given to tracking activities and cross-referencing all available information. Another possible objection is that such tracking activity constitutes an excessive form of control over the user and those who preceded them. However, the intention here is not to monitor anyone’s private life, but simply to keep track of the activity of an operator and their predecessors \textit{within the strict boundaries of that activity}, which may carry significant weight in terms of \textit{risks to others}. Not to mention the fact that an artificial intelligence system is always integrated within a broader context: for example, a physician primarily observes the patient with their own eyes and touches them with their own hands, often accompanied by a team. Another issue may arise from the use of large datasets, a rather concrete matter, perhaps more ethical than legal. Indeed, even if one were to place the responsibility on the dataset's creator, from whom biases might arise, for example, in Step 3 of our protocol, it is nonetheless practically difficult to correctly manage large volumes of data. Once again, it is a matter of balance. 

A positive note might instead regard the recital 53 of the AI Act \cite[pp. 14-15]{EU2024AIAct}, concerning \guillemotleft specific cases in which AI systems referred to in pre-defined areas specified in this Regulation do not lead to a significant risk of harm to the legal interests protected under those areas\guillemotright. Among the conditions that should be fulfilled in order for a system to take a step down in the pyramid of risk, for example, the second one states that \guillemotleft the task performed by the AI system is intended to improve the result of a previously completed human activity that may be relevant for the purposes of the high-risk uses [...]. Considering those characteristics, the AI system provides only an additional layer to a human activity with consequently lowered risk\guillemotright. We believe that the Endless Tuning could align at least with such requirement, perhaps entailing not insignificant advantages also from an economic viewpoint.

\section{Conclusion}
\label{sec:chap6}

Do you remember the young Fred? Well, after our experiments, we might justifiably assert that, by following the principles of Endless Tuning, not only would John, the doctor, not have resigned, but the administration would have had no reason to place blind trust in the system. Instead, the skills of both the doctor and artificial intelligence would have worked in synergy. In this paper, we provided some reasons for a \textquotedblleft different voice'' in artificial intelligence ethics, step by step illustrating a protocol, implementing three prototypical versions thereof and testing them with as many domain experts. Our goal has been to avoid user replacement and to give the possibility of tracing back responsibilities. Considering the reported user experience, a relational (or hermeneutic) approach seems to hold great promise, even though, as emphasized in various sections of this article, every idea has, or may have, its limitations. Similarly, such an approach was proposed according to the best of our knowledge. There are, as always, many potential uses for an idea, and we ourselves could have explored the Endless Tuning, for example, for educational purposes. Some limitations may also be found in the very experimental process, since a complete test in the long run would have been required, yet entailing too much time and, perhaps, organizational costs. Nevertheless, this means that there is room left for future research --- even sincerely hoping that the Endless Tuning design could serve as inspiration and bring benefits in the field of generative artificial intelligence too, allowing people to participate right in the generation process. 

This is but a single drop, yet countless drops have the power to form an ocean.

\section{Acknowledgment \& Declarations}
\label{sec:chap7}

Work supported by PNRR - M4C2 - Investimento 1.3, Partenariato Esteso PE00000013 - \textquotedblleft FAIR - Future Artificial Intelligence Research'' - Spoke 1 \textquotedblleft Human-centered AI'', funded by the European Commission under the NextGeneration EU programme. We also would like to thank the members of the Facultad de Filosofía y Letras of the University of Valladolid, Spain, who supported the writing of this article making us warmly received during a research stay and suggesting us valuable insights.

We declare that the people involved in the experiments agree with the publication of results and transcriptions, and also that the author has no conflict of interest.

\bibliographystyle{plain}  
\bibliography{bibliography}

\appendix
\newpage
\section{Transcription of the bank director's interview}
\label{AppendixA}

Age of the interviewee: 55

Work experience in a bank: 30 years

Work experience as bank director: 24 years\newline

I: So, the idea is this: the machine in question is a decision tree that was trained on 20,000 cases from a real, anonymized U.S. dataset. You’ll see in a moment that there are various parameters --- it’ll all show up in the table shortly. It’s called Endless Tuning because it works like a two-way mirror. The idea is that you can give the model some kind of learning input as you go, and in return, the model gives you suggestions each session to help you make better decisions. Over time, obviously, it’ll sort of adapt to your style... So, [looking at the initial screen] this part we’ve already read, no problem. “I understand.” [The director clicks on the “I understand” button.]
Of course, you are an actual bank manager, and someone is now requesting a loan from you.

BD: [Murmuring while reading the table…] Wait. We need to decide whether...

I: Whether it’s worth approving the loan or denying it to this person.

BD: Okay.

I: Based on your experience --- and only if you want to --- you can also rely on my AI system, but you're not required to. At this point, we need to pick a number at random, between 0 and 200. That would be the ID.

BD: Client X.

I: Exactly. Shall we pick...?

BD: 144.

I: So, here are this person’s details. We can also write some notes --- it would actually be helpful, to better understand things.

BD: [Speaking to himself] 28 years old, male, bachelor’s degree, income... well, dollars or euros, doesn't really matter here. He’s asking for 20,000 dollars for personal reasons.

I: The interest rate is $ 10.99\% $ [director’s surprised look]: yeah, pretty high. And this is the loan amount as a percentage of the annual income --- $ 27\%$. “Credit history” means how many years in his life he’s had loans, etc.

BD: “Credit score” is a number that ranges from...?

I: The American average is 705. It’s the probability...

BD: ...that the loan will be paid back. The higher, the better. And he’s never defaulted. Okay, should I say if I need more information?

I: Not quite. Right now, the first point is: what do you think?

BD: Worth considering.

I: Great, we can go ahead and write “worth considering.” Is there anything in particular about these details that stands out to you?

BD: Maybe the years of employment. Not too few, but not that many either. What we’d usually do [referring to his potential course of action]… well, this might be something specific to the Italian job market, because in the U.S., the reliability of the employer... over there, the ability to move from one to another is quite, let’s say… [he means: easy]. Whereas we tend to check whether the employer... we’d do a couple of basic checks: see if there are any serious red flags, past defaults, or if it’s a company that was founded, let’s say, two months ago --- maybe three years ago --- so, if it has little experience. We also look at the employer’s general solvency.

I: So [writes down the note] “we would also assess the employer’s solvency,” which is missing here as a parameter.

BD: Much less so, though --- just very lightly. We’d look only for serious issues: liquidation, ongoing bankruptcy, because sometimes the company might be on the brink of collapse.

I: An attribute that, in this case, you don’t have access to in the program, but you obviously know to consider. So, your initial instinct --- approve or deny?

BD: Yes, yes, I’d approve it.

I: Perfect, let’s move forward. Would you like to use the AI? Or should we just skip to the next one?

BD: But what would I see?

I: You’d see a suggestion.

BD: I’d do it out of curiosity, sure.

I: Great. So what’s the idea here? “Keep in mind that…” This [pointing to the right side of the screen] is our case, of course. “Keep in mind that” he has no defaults, and what he’s asking for is more than $ 23\%$ of his income...

BD: Less.

I: No, more --- it’s 0.27.

BD: Oh right --- our bank’s threshold is $33–35\%$, that’s why I got confused.

I: You have a reference point, so you tend to... [the director nods---it’s understood: he tends to align with that reference]. He’s renting, yes, and then “percent income,” yes --- it’s also higher than 0.24. We’ll have to see how the tree splits here, because we can’t tell right away. Would you like to add any notes? Has your opinion changed in any way?

BD: Ah, one thing --- the rent: it’s not specified, right? Because having it quantified would help you understand, for example, that the fixed monthly expense he has can’t really be reduced or adjusted. Like, utility bills aren’t counted here because they’re variable --- if you use less electricity, you pay less. But rent, on the other hand, is a non-negotiable expense.

I: [Helps writing] “Rent amount is not specified here.” Moving on. At this point, you have a second clue, which consists of this: it evaluates similarity --- based on compressed representations, so there’s some margin for error in the interpretation --- with the three most similar past cases. Important to note: in these cases, the “target” label --- whether the loan was granted or not --- was assigned by whoever previously labeled the data. That person may be more experienced than you --- or less. Naturally, there’s no absolute standard here; it depends on many factors.
In addition to that, you also have a graph to get a general sense of the positioning [relative position]. All three, by the way, were denied.

BD: Wait, I didn’t catch that. These cases are...?

I: The most similar ones.

BD: Ah, of course… by age, different gender (but that doesn’t matter), level of education… 12 years, 0, 5, 3. Renting, here there’s a mortgage, here it’s ownership and rent. This is the loan amount.

I: This one asks for \$9,000 to consolidate debt, and this one here, a mortgage of \$8,000 for school or university, basically.

BD: So this person has a 10-year credit history in the databases --- what does that mean?

I: Credit history means how many years they’ve spent with loans, financial obligations, applying for credit, etc.

BD: Hmm, it seems very restrictive to me.

I: It’s totally possible they got it wrong.

BD: Especially in one case --- this one, number 2. They had defaults, so it makes sense. We’ve seen it has a lot of weight, because this data doesn’t show up indefinitely. We only see it for 2–3 years, and if it appears, it becomes much more problematic.

I: [Helps write] Seems like a restrictive policy, especially regarding the second column.

BD: Here, maybe what influenced it is this --- this person is newly hired, right? But they earn \$104,000. That’s why I’m saying, the employer matters. If they work for a major microelectronics company, or if they’re employed by a little restaurant around the corner that’s only been around for two years...
They’ve got a mortgage, but we also need to see what other financial obligations they already have. That wasn’t included in the original data either. There’s no indication of other existing commitments with the banking system. And whether that percentage in relation to income...
We consider not just $33\%$ of our payment, but the total debt burden the client carries. I mean, we check the credit databases, where all institutions report. If in addition to paying us €300, they’re also paying €1,000 to other banks, that percentage isn’t 0.26 anymore. It becomes 0.50.

I: [Helps write] Are you still leaning toward approving it?

BD: The mortgage… that piece of info is important.

I: In the end, you’ll still have the option to pause your decision. For now, would you say you’re leaning more positive or more negative?

BD: Positive.

I: Now we’ll get the actual output, let’s say, the system’s final decision. The tree was completely in favor, with 99\% confidence. These are your previous notes. Of course, you can still choose not to proceed --- you can decide to deny or approve, for various reasons. You’re not fully confident? No problem.

BD: It’s the information on existing obligations that’s decisive.
[The interviewer helps transcribe the note.]

I: Excellent. Due to time constraints, we won’t go through more cycles now. Gradually, once you apply your label, you’re building a new, ever-growing dataset. Once that reaches a sufficient size to be accurate, the model is retrained the model --- then one can decide how flexibly to do that, whether to mix in earlier data or not.
However, if the first key feature was that you could get some support from the system, the second is that there’s an issue of accountability. Suppose something goes wrong. In that case, there’s a log available --- of course, here it’s just a regular .csv file, but the idea is more like an immutable black box, as in aviation. [Shows the log.] This is exactly everything you’ve done. On April 16th, 2025, you loaded case 144, etc.

BD: The basic idea is that there’s a trace left by the system.

I: Exactly --- there’s a trace. Since everything random is saved (and everything else is implicitly deterministic), you can replay the whole process, step by step. So, for instance, it would be possible to determine whether something in the interface design might have misled you. Suppose I wrote all the suggestions in white, except one in red.
Now, if I retrained the model with no safeguards, I could end up with very poor performance --- and that would be on me.

BD: But could there be implicit correlations? For example, you’d need to know the monthly payments of a loan to understand what some factors really mean.

I: That’s a fundamental challenge: on one hand, AI only sees what it’s fed; on the other hand, data can carry meanings that aren’t immediately obvious --- and there could be proxy attributes.

Now I’d like to move on to a few questions.

BD: Of course.

I: Did you find the model’s suggestions useful? If so, which ones?

BD: The usefulness of the first suggestion was mainly in comparing thresholds. We, for example, use a 33\% threshold.

I: And what about the similarity-based suggestion?

BD: No, not really, because I evaluate each case individually. Using the same parameters I looked at earlier, I would’ve examined all four cases. Where there was, say, a history of default --- in the second one [second column, which was originally denied] --- I might’ve approved it anyway. But not because I compared it to the others, just because of certain specific features.

I: If you had to rate this type of interface from 0 to 5?

BD: 4 --- because there were those two key pieces of information that were missing.

I: Did you ever feel, in any way, replaced by the intelligent system?

BD: No.

I: Would you consider this a reliable approach? I mean in two senses: first, the reliability of the system itself; second, the reliability of the approach --- where essentially you are the first to decide, and over time... there’s no program giving you an answer before you express your own judgment; it only gives feedback at the end.

BD: Reliability... my impression was that the decision-making part, honestly, felt more mine than the machine’s.

I: Exactly: the machine provided everything it could --- what led to the decision, the similarity data, and then the final outcome. The key point is that the machine didn’t immediately say: “approve” or “deny.”

BD: But that thing at the end, that final output --- was that a calculation the machine made based on my observations?

I: No, no --- it was a calculation the machine made on the case itself. More precisely, if we really want to go into detail, the machine had already made that calculation at the beginning, but didn’t reveal it. It was only shown at the end. In future sessions...

BD: So that final result could have actually contradicted my choice?

I: Of course --- it’s independent, at least for the individual decisions. Over time, it will update.

BD: In that case, based on direct experience, I’d say it’s reliable. From this perspective, it reached the same conclusion I did by reviewing things...

I: The model was accurate on that dataset at 83\%.

BD: So, all in all, fairly solid. If only that one additional piece of data had been there...

I: That’s right---those data were missing.

BD: External obligations.

I: And finally, any general thoughts on the decision-making experience? Not so much about the machine itself, but more about your experience with this kind of program?

BD: I’d say it was positive. Because, again, having summarized data and having, let’s say, some additional reassurance that aligns with my own perception --- based on my experience --- and having a confirmation of what I’m seeing helps reduce the margin of error. Because human error --- say I read a figure as 27\% when it was actually 37\% --- that can happen. So, if the machine is well calibrated, and at the end says, “Look, in my view this is a no, but…” --- that’s helpful.

I: One more question: if the result had contradicted your opinion, would that still have helped you? Or would you have just said: “Nope, it doesn’t agree with me, so it’s useless”?

BD: In that case, I’d go back through the process and try to understand what part diverged from my assessment. As a human, if I make a mistake, understanding where I went wrong helps me in the future. If a tool gives me that kind of feedback afterward, I’d say --- it’s useful.

I: Very good, that concludes our session. Thank you very much.

\newpage
\section{Transcription of the aesthetics professor's interview}
\label{AppendixB}

Age of the interviewee: 46

Experience: 30 years\newline

I: Okay, so the Endless Tuning is a design method of artificial intelligence, a relational design method.
It aims at two goals. The first one is to use, at least in decision-making processes, artificial intelligence without replacing the user and having the possibility to trace back responsibilities.

PAE: Okay.

I: The two great problems.

PAE: Okay.

I: So it aims at finding a solution within a double loop, two negative feedbacks, one in front of the
other. The idea is that you can put constraints to the model, learning constraints, and the model will be
able to let you reflect about your single-case decisions. The first one is made session by session by
building up a new dataset based also on your labels and after a certain number of sessions reached,
(which was simple one in this case, just to see it), it retrains.

PAE: Okay.

I: So, it adjusts itself to the local environment.

PAE: Okay.

I: In the second place, you will have some saliency maps, of the single case that we will treat, and you
will have a measure of similarity with similar cases.

PAE: Mm-hmm.

I: If you want, you can start.

PAE Okay. Let's go. Let's go. Ha! I'm going to test an application of AI. Yes. Yes. And I'm very happy of it
[clicks on \textquotedblleft I understand''].

I: Okay. So, you can choose. We have this dataset and we have seven possible classes. Obviously, I
can't tell you which are the classes. If you want, just some folders, one for classes… Yes, choose one if
you want.

PAE: Uh-huh.

I: What do you prefer?

PAE: Between these two? Uh-huh.

I: This is for the class. This is for the single image. Just choose one.

PAE: There's one between these six possibilities?

I: Seven.

PAE: Seven. Uhm... The classes [The interviewee appears not yet fully aware of how to carry out the
task].

I: When you want, you can also choose the image within a class, by selecting the image. What do you
like?

PAE: I don't know. I think that I like this style.

I: If you want to change image, among the same style, you can just choose, select an image.

PAE: Other possibilities or the same classes. So really? I don't understand what I am doing.

I: You're just choosing a single case.

PAE: Okay. A single case between these possibilities.

I: Yes. To classify.

PAE: [Now suddenly aware] Ah, okay! These are classes! Okay. These are classes. Okay. Okay. Okay.

So, I'm going to choose this one. And then we have... [class] 1, [case n.] 1495. Yes. Class 1, 1495.
[clicks on \textquotedblleft Proceed'']

I: So, we have to recognize the style of this artwork. The idea is, first of all, which is your impression? If
you want, or it might be better, you can add some notes just to take notes of your work.

PAE: My personal impression.

I: Yes, first of all, completely free.

PAE: Completely free of mine. Okay.

I: And then you will have to select a possible class in 
the end, in the end of this session.

PAE: Okay. Among these same ones.

I: If you are not sure, don't worry. Just put the class that fits best to you. Because in the end, you will be
given the possibility to \textquotedblleft I will choose later’’. You can put everything in just your
language and...

PAE: In my language? Well, I'll be able to use my language. Um... Any note?

I: Any note.

PAE: [Writes] «I am especially impressed by the upward gaze toward the mystical heights that seem to
illuminate, in return, the main subject. Likewise, the overhead lighting and the attitude of serenity and
contemplation are captivating». So, \textquotedblleft select a class''... These classes are artistic styles, not aesthetic
attitudes, right?

I: Yes. Art genre.
[The interviewee selects the class \textquotedblleft Baroque'']

I: Now, we have a first clue. It will take a few moments. The first clue is simply clouds of importance on
these images. The idea is simply: «Look here». It is meant to give you simply a suggestion. Technically
speaking, it is an explanation of our model since our model already decided, but we don't know what,
also because we don't know if the model is right or wrong. So, this is an explanation of the model that
I'm using in a reversal mode as a hermeneutic process. [Processing is done] Obviously, yeah, it might
have paid attention to irrelevant factors.

PAE: So, okay. Wow.

I: From blue to red, importance is increasing. It's just: «Consider also this». Obviously, you can still
provide some considerations if you change idea. You can confirm, you can change class.

PAE: This saliency map is done from my prompt, my description?

I: No, no. This is a suggestion. You're directly hearing, say, \textquotedblleft the voice of AI'' in this moment.

PAE: So, I'm discussing with it.

I: Exactly.

PAE: I partially agree with it.

I: It may be wrong too, obviously. The accuracy of the model was 65\% so it's possible to be wrong.

PAE: I'm right with this, uh… this analysis, this zone from here, yeah, but I think that is very important,
the rest of the body. I think, uh…

I: If you want you can write it.

PAE: [Writes, referring to the RISE suggestion] «I believe there are some issues with the evaluation of
the image. I agree with the idea that the face and upper body posture are important, but the lighting of
the whole body is crucial, partly because it contrasts vividly with the background». Perfect.

I: You confirm, \textquotedblleft Baroque''?

PAE: Ok. Confirm.

I: And it's another point. You see why? Let's say issues in AI. We have the same black box, but two
different explanations. Because we used two algorithms. However, the difference is that before we
used an algorithm that was mounted on the entire black box.

PAE: Ok.

I: So, we just focused on clouds. Here, we are focusing just on one layer that is able to perceive single
characteristics.

PAE: Single characteristics. So it's more detailed.

I: This single section of suggestions, of clues, is focused much more on details. But the significance
colormap is the same. Red is more important. And it is just meant to say to you: «Look also here». If
you want, obviously you can enlarge the image.

PAE: No, no, I know the image. So... I'm not sure about this. No. Perfect. No. Because I see the sense of
focus, because in fact there are a lot of details in these signs but they are not significant. I’m thinking
to the whole of the image. So, I can write... This is very interesting. It is fun.

I: You will see why I'm asking you to write in the end.

PAE: [Writes] «I disagree with the assessment because, even though the focus on the level of details is
understandable, this fixation is merely formal. Since the details are not particularly significant for the
overall meaning of the image». Ok. I believe yes. Confirm [confirms \textquotedblleft Baroque''].

I: Now... While training this model, obviously I have to train it on some images. According to a
compressed representation (so please note that compression might induce some error), I'm ranking
the most similar images to the present case. And you also have a diagram here, just to simply give you
a relative distance of the cases. So... The idea is just comparing them, you might have also some more
idea, you can change the idea, and so on.

PAE: Uh-huh. Mmm... Mmm... That is, the closest to the image. Mm-hmm. But it's... According to its
compression?

I: According to all the images. All the images, comprised this one, were compressed, and then
distance, geometrical distance was measured.

PAE: I think that's... I don't know really, I don't know. I believe that from a point of view, from a formal
point of view, I think that it's... It's right, but according to... According to a content point of view,
content, sense, meanings... Yes. Really, I don't know exactly if... Mmm... So what can I do?

I: Obviously the idea is just to give you some suggestions, so you can still write words, you know, then
choose another class and so on. You can do whatever you want.

PAE: Okay, well, I'll put my reflection on the particular, which is... [writes] «From the formal point of view, I understand that it's closer to the head of Medusa, because the drawing and the patterns are actually
similar. But from the material point of view, I think they're not so close». I don't know if I... I'm doing right.

I: It’s okay. Do you want to choose another class, or you feel you can confirm Baroque?

PAE: No, I confirm Baroque. Proceed.

I: Now, you will literally hear the voice, \textquotedblleft the final voice of the AI'', what we didn't hear.

PAE: So, what do we have here?

I: Here is a histogram with results, together with confidence for each class.

PAE: Yeah. Predicted by AI.

I: In this case, the AI is strongly agreeing with you. You have in this case, 95\% sure about saying
Baroque. And so simply, if we before received just some suggestions, now, we can really talk with the
model. Anyway, note that the system is modular, so I can exclude each module if I want. It just
depends on the environment. What I'm giving you here is, in the extreme case, a comparison.

PAE: Mm-hmm. Okay. Confidence table.

I: So, I think we could confirm.

PAE: Yeah, confirm.

I: So, I will give you an insight. What will happen, since I just chose 1 as the number of required
sessions, is retraining. While doing training, the original one, I saved a portion of data. I take
a piece of this data, a little piece for each class, except the one that you chose. So, when we will click
on \textquotedblleft Proceed'', \textquotedblleft Baroque” will be given as label to this image. This image will be put alongside other
images from the other classes. And a new data set with your instructions will be built, little by little.
Okay. So, after a certain number of sessions, my model will be able to retrain itself, but with global
suggestions [the model retrains]. So, why did I ask you of writing? Let us suppose that someone wants
you to account for what you did. Note that we have a double problem here. One is \textit{ex ante}
accountability, the other is \textit{ex post} liability. It is not so easy to trace back responsibilities. [Showing the
logs]. We can record. We can know the maps which were randomly produced. Since the model is
intrinsically deterministic, once we add the image, it will give always the same result. And we have
each minute and second here... everything you did. Even that the model parameters have been tuned.
It means that... You tune with the model in a certain sense.

PAE It's a sort of... coincidence.

I: This is the endless tuning. And a judge will be able to review the entire process is slow motion.

PAE: Okay, I understand. This is really... this is very important. Yeah, this is very important.

I: So, in the end, well, I just need to ask you just four questions. So, first of all, did you find useful the
suggestion given by the saliency maps?

PAE: Yes. Yes. Hell yes. Maybe not for me exclusively because I am an expert in this field. But I think
that they are very useful. Because it's a copilot for discussing with it. Same thing when I translate from
Latin, for example, and I use LLMs and I'm translating like ever, like always. And I say, okay, LLM, what
do you think about this translation? And it gives some kind of output. I think I can choose more
efficiently and I am more confident with this class. But because I know translating from Latin, I know
history of art. So, for me, it's useful. And I think that it can be useful for anyone that doesn't know
about… It's not so explicit. I think this is very useful.

I: What do you think about the similarity comparison as a process? Might it be useful or was it useful
for you?

PAE: Not so much. For me, it's been… No, yes, but it's been surprising. 

I: Surprising? 

PAE: Yeah. Because I didn't understand exactly why this… those images… Mystery [laughing]. Yeah. I have… I've understood the second explanation because I'm not okay with it, but I understand it.

I: And did you feel being replaced by the AI?

PAE: No. I've been feeling all the control alongside the process.

I: Final question. From 0 to 5, which vote would you give to the interface? Not much the accuracy of the
model, which obviously is a toy one.

PAE: Yeah. But of the… system's interface. Which keys for it? Accessibility, design or…

I: I would say accessibility. What we are evaluating is the design, the process, the loop, on the whole.

PAE: I think that it's very clear, very simple. So, I give… just as an incentive, 4 out of 5.

I: Thank you.

\newpage
\section{Transcription of the doctor's interview}
\label{AppendixC}

Age of the interviewee: 59

Experience: 27 years\newline

I: Let’s randomly pick a case from a class.

R: 1197.

I: Excellent. So, the first step is simply to give your initial impression. The task is binary: just say whether it has pneumonia or not.

R: We need to zoom in… [zooms in on the image]. Look, there’s only one point of concern here, but it’s probably just a calcification of the sternocostal joint. If I had the chance to adjust the windowing...

I: You're right, but this is just a toy example. As we mentioned earlier, the experiment is focused on the interface. Would you be so kind as to write down what you just told me as a note? We’ll see why later.

R: [Writing] Single concern in the lower right field, likely an artefactual image. That’s what I would say.

I: Perfect. So, as a first impression---does this case have pneumonia?

R: No, to me there’s nothing.

I: Let’s move on. Now this will take a couple of minutes to process. What happens next is that, using an algorithm called RISE, which operates over the entire network, we’ll try to extract in real time some clouds of possible importance on the image surface.

R: You mean areas?

I: Yes, areas. Specifically, the redder ones could be more important; the bluer ones, less so.

R: A question: how does the program even know there are cases with pneumonia? Someone must have told it?

I: I had data compiled by people who are presumably experts.

R: Got it.

I: But that’s one of the risks and weaknesses of artificial intelligence: it requires clean data.

R: What do you mean by \textquotedblleft clean data''?

I: \textquotedblleft Clean data'' means, for example, data that is evenly distributed among classes, or clear enough, with minimal information loss. This is a whole science in itself. But now the processing is complete. What we see here is a map overlaid on the image of our study case --- which remains here on the right in its original form. The system suggests that these two points might hold greater relevance. Keep in mind, as noted, the map might not be entirely reliable --- and more importantly, our model may have focused on irrelevant features.

R: Can we look at it again? This map definitely points to incorrect features. No, I’m not convinced. In fact, [in those points] there’s nothing.

I: There’s nothing.

R: I disagree. [Writes] I confirm the hypothesis of no pneumonia.

I: Now here’s another clue. On the same black-box model and the same case, we’ve applied another algorithm called Grad-CAM. The difference compared to RISE is that while RISE analyzes the entire network (or more precisely, the output of the network through a different process), Grad-CAM looks at a specific layer. Since this is a convolutional network, it focuses on a layer that should reflect more detailed features. So this map doesn’t highlight broad areas, but more specific points that might be worth considering. The color scale follows the same principle.

R: No, I still maintain that there’s nothing. I don’t agree with these suggestions.

I: Now let’s introduce another clue. What’s the idea here? So far, we’ve been “talking” with our model via algorithms --- explainers --- applied on top of it. But now, we take all the training images --- around 2,500 --- which have about 85\% accuracy overall. These were annotated beforehand by doctors, who may be more or less experienced than you. This time, it’s as if you’re speaking with those annotators. We’ve compressed these images so we can plot and compare their relative proximity to the case at hand. The three most similar images --- according to the compressed representations --- are shown. Of course, compression entails some loss of information, so the similarity may be questionable.

R: Here it says none of the three has pneumonia. So this is according to the data annotators, correct? Alright. [The comparison, even if not strongly supporting the initial hypothesis, is at least not contradicting it, while the user’s confidence seems to increase as they proceed.]

I: Exactly. So there’s, let’s say, agreement between you and the annotated data. Not much to add then --- we can move on. This is the actual decision of our neural network, expressed not in binary terms but as a confidence histogram. In this case, the model says there is no pneumonia with 80\% probability. So, in fact, you’re also in agreement. Note that we highlight the possibility the model may not be well-calibrated --- we call that being overconfident. So, do we confirm?

R: Yes, we confirm that there’s no pneumonia.

I: What happens next --- after this single session, for demonstration purposes --- is that the model will be retrained. I had set aside a small reserve of data, and now, using the label you just provided, I’ll begin constructing a new dataset to be fine-tuned later. The underlying idea is that the model moves closer to us and our operational context --- but does so by prompting us to reflect, by offering hints and suggestions.

R: I understand completely.

I: Now, let me explain why I asked you to write a note earlier. Suppose that following this decision, some harm occurs and someone files a complaint. Even in this basic, illustrative implementation, a real-time log is compiled, recording every interaction with the system. We know the model and its weights --- it is deterministic. We have the original data, the saved maps, and the result will always be the same given the same input. So, if this were taken to court --- despite the fact that AI models today are recognized as having some degree of autonomy and unpredictability --- the judge could review everything like a replay. The idea is that, rather than pre-assigning blame, we can use an informatic component to trace the distribution of responsibilities.

R: But you mean the responsibility could lie with the doctor or the AI?

I: It could lie with many more parties: the model developer, the interface designer, the user --- whose actions are all logged --- and in extreme cases, it could be a matter of chance.

R: Wait, let me understand. I think there are two situations: one in which I agree with the AI and we both get it wrong. In that case, the judge might say: “Excuse me, why did you believe the AI when it was clearly wrong?”; and another situation in which I disagree with the AI --- the AI says one thing, and maybe it’s right. If I trust the AI, can I avoid professional liability by saying, “Well, the AI told me”? And if I don’t trust it and I’m wrong, can I defend myself by saying that I had the right not to believe it --- and that this shouldn’t count against me? And if the AI steered me toward an incorrect diagnosis, does that excuse me, or not?

I: That’s exactly the issue. The goal here is to avoid total exemption from responsibility. That’s why the interface provides hints --- to get the user thinking and thus sharing some responsibility. This doesn’t mean you can’t agree with it, of course. But there’s something called automation bias --- we tend to trust machines too much.

R: In my opinion, it’s the opposite… but anyway, go on.

I: So then, who’s liable? The computer isn’t intelligent but it behaves --- the risk is that damages go unpaid, or that blame is arbitrarily assigned to one party.

R: I think ----- though I’m not a judge ----- that the judge, with their culture and human sensitivity, and with zero awareness of how AI truly works, has to decide whether the AI said something foolish or not.

I: Well, as things stand, it’s a human judge who must resolve disputes.

R: But perhaps this should be standardized.

I: The European Union is aiming in that direction, indeed. But these are still very complex technologies, and standardization is difficult. Now, just a couple of questions. First: your overall user experience score, from 0 to 5?

R: Honestly, I wouldn’t know how to rate it ----- this is my first experience testing an AI system, and I wouldn’t know how to weigh the pros and cons.

I: And regarding the suggestions?

R: The maps were completely inaccurate: what the system flagged was completely unreasonable --- it highlighted features that couldn’t possibly mean anything.

I: Lastly, did you feel in any way replaced by the computer?

R: Absolutely not.

\end{document}